\documentclass[11pt]{article}

\usepackage{microtype} % microtypography
\usepackage{booktabs}  % tables

\usepackage{amsmath}
\usepackage{amsthm}

\usepackage[final]{automl}

\usepackage{natbib}
\def\nfo{DIFER}
\usepackage{graphicx}
\usepackage{caption}
\usepackage{filecontents}

\title{\nfo{}: Differentiable Automated Feature Engineering}

\author[1]{\nameemail{Guanghui Zhu}{zgh@nju.edu.cn}}
\author[1]{\nameemail{Zhuoer Xu}{zhuoer.xu@smail.nju.edu.cn}}
\author[1]{\nameemail{Chunfeng Yuan}{cfyuan@nju.edu.cn}}
\author[1]{\nameemail{Yihua Huang}{yhuang@nju.edu.cn}}

\affil[1]{State Key Laboratory for Novel Software Technology, Nanjing University}
\hypersetup{%
  pdfauthor={AutoML-Conf}, % will be reset to "Anonymous" unless the "final" package option is given
  pdftitle={Documentation for the automl Package},
  pdfsubject={Documentation for automl Package},
  pdfkeywords={AutoML-Conf, LaTeX, style}
}

\begin{document}

\maketitle

\begin{abstract}
Feature engineering, a crucial step of machine learning, aims to construct useful features from raw data to improve model performance.
In recent years, great efforts have been devoted to Automated Feature Engineering (AutoFE) to replace expensive human labor.
However, all existing methods treat AutoFE as an optimization problem over a discrete feature space, leading to the problems of feature explosion and computational inefficiency.
Unlike previous work, we perform AutoFE in a continuous vector space and propose a differentiable method called \nfo{} in this paper.
Specifically, we first propose an evolutionary framework to search for better features iteratively.
In each feature evolution step, we introduce a feature optimizer based on the encoder-predictor-decoder, which maps features into the continuous vector space via the encoder, optimizes the embedding along the gradient direction induced by the predictor, and recovers better features from the optimized embedding by the decoder.
Extensive experiments on classification and regression datasets demonstrate that \nfo{} can significantly outperform the state-of-the-art AutoFE method in terms of both model performance and computational efficiency.
The implementation of \nfo{} is avaialable on \url{https://github.com/PasaLab/DIFER}.
\end{abstract}

\section{Introduction}

% background: introduce resource-intensive feature engineering. 强调人力过程
% More data beats clever algorithms, but better data beats more data. 
Feature engineering, the process of constructing features from raw data, directly determines the upper bound of various machine learning algorithms (e.g., Random Forest and Logistic Regression).
However, it requires considerable domain knowledge to construct features.
Also, huge computational resources are needed to evaluate and then filter features.
Thus, it is a cost-intensive task to find useful and meaningful features.

Recently, the AutoFE (Automated Feature Engineering) methods that search for useful features without any human intervention have received more and more attention.
AutoFE formalizes feature construction as applying transformations (e.g., arithmetic operators) to the raw features.
The \textit{expansion-reduction} algorithm \citep{kanter2015deep,lam2017one} iteratively applies all transformations to each feature and selects
the features based on the model performance.
Without expert guidance, such method consumes significant computational resources for feature evaluation due to the exponentially growing feature space.
To reduce the cost, learning-based AutoFE methods are proposed.
% TransGraph
TransGraph \citep{khurana2018feature} trains a Q-learning agent to decide the transformation.
Due to applying each action (i.e., transformation) to all features, TransGraph also suffers from the feature explosion problem.
% LFE
LFE \citep{nargesian2017learning} trains an MLP (Multi-Layer Perceptron) to recommend the most likely useful transformation for each feature.
However, it does not support the composition of transformations.
% NFS
NFS \citep{chen2019neural} generates a feature transformation sequence for each raw feature under the guidance of an RNN controller.
% The controller is trained with policy gradient \citep{sutton2000policy}, which views the performance of transformation sequences and their sub-sequences as the reward.
Although NFS can achieve SOTA (state-of-the-art) performance, the computational efficiency is still low.
%
% Problem of Automated FE
An inherent cause of inefficiency for the existing approaches is the fact that AutoFE is treated as an optimization problem over a discrete space.

% motivation & method
%Inspired by the recent work DARTS \citep{darts} and NAO \citep{nao} for differentiable neural architecture search, 
In this paper, we address the AutoFE problem from a different perspective and propose the first gradient-based approach called \nfo{} (DIfferentiable automated Feature EngineeRing).
% Instead of searching directly in the discrete space, we encode features into a continuous vector space and perform gradient-based optimization in the continuous space
We first propose an evolutionary framework to generate better features iteratively.
Then, in each feature evolution step, we propose a tree-like structure called \emph{parse tree} to represent constructed features flexibly, and leverage a feature optimizer based on the encoder-predictor-decoder. 
Specifically, instead of searching in the discrete feature space, the encoder maps the traversal string of the parse tree into a continuous vector space.
Constructing a better feature is equivalent to generating better embedding in the continuous vector space.
The following predictor takes the feature embedding as input, predicts its performance score, and directly optimizes the embedding by gradient ascent along the score direction.
The optimized embedding is further decoded as a better feature in the discrete space.
%% 
%Furthermore, based on the feature optimizer, we employ a feature evolution method to search for better features iteratively.
% Moreover, \nfo{} is much more fine-grained to perform transformations on each raw feature rather than the entire dataset.

%% simple intro
%Specifically, we leverage an encoder-predictor-decoder framework as the feature .
%Similar to the seq-to-seq learning \citep{sutskever2014sequence}, we use the encoder and the decoder to learn the bijection between the discrete feature space and the continuous vector space.
%Thus, extracting better features is equivalent to generating better embedding in the continuous vector space.
%% execute the algorithm on such a massive network for a large number of times
%Then, the predictor takes the feature embedding as input, predicts its score, and directly optimizes the embedding multiple times by gradient ascent to get a better feature after decoding.

% experiment info
Extensive experimental results on both classification and regression tasks reveal that \nfo{} is not only effective but also efficient.
Compared to the SOTA approach, \nfo{} achieves better performance on 22 out of 25 datasets with 40 times fewer feature evaluations.
Moreover, \nfo{} can be effective when using different machine learning algorithms.
%%%
%%Compared to the SOTA AutoFE approach \citep{chen2019neural}, \nfo{} achieves better performance on 20 out of 23 datasets with much fewer feature evaluations.
%%The experimental results also demonstrate that \nfo{} is independent of machine learning algorithms.
%%Furthermore, the experimental results with different maximum orders of transformations prove that the performance of \nfo{} continues to increase with the order of feature space.

% summary
To summarize, our main contributions can be highlighted as follows:
\begin{itemize}[topsep=0.4ex,parsep=0ex]
\item We propose a feature evolution framework to search for better features iteratively.
\item To represent constructed features, we design the \emph{parse tree} structure, which is more flexible and expressive than the commonly-used sequence representation.
\item We introduce a novel feature optimizer based on the encoder-predictor-decoder for feature evolution and thus can achieve differentiable AutoFE.
To our best knowledge, \nfo{} is the first differentiable AutoFE method.
\item Extensive experimental results on a variety of tasks demonstrate that \nfo{} outperforms the state-of-the-art AutoFE approach in terms of both model performance and computational efficiency.
\end{itemize}

\section{Related work}

% 大致介绍自动化特征工程的分类
Feature engineering aims to transform raw data into features that can better express the nature of the problem.
%Training models with generated features can improve prediction performance.
Recently, feature engineering has gradually shifted from leveraging human knowledge to automated methods.
Existing AutoFE approaches can be divided into three categories.
%The main challenge of AutoFE is the contradiction between limited resources and the considerable feature space.
%Each time a new feature is extracted, a machine learner needs to be trained from scratch to accurately evaluate its performance.
%As features are constructed in an alternating multi-step process by applying transformations to the features, it results in an exponentially growing feature space.
%To resolve this contradiction, efficient search strategies or low-calculation surrogate criteria are needed urgently.
% \citep{bagallo1990boolean,markovitch2002feature,cheng2011automated,kanter2015deep,nargesian2017learning,khurana2018feature,chen2019neural}.
% As mentioned above, AutoFE has gone through the development process of surrogate-criteria, knowledge-based, expansion-reduction, and learning-based stages.

%% surrogate-criteria
%\paragraph{Surrogate Criteria.} Rather than optimizing for the prediction performance directly, FICUS \citep{markovitch2002feature} evaluates new features generated by beam search based on surrogate measures such as the information gain in a decision tree.
%Similarly, FCTree \citep{fan2010generalized} also uses tree-based information-theoretic criteria to guide the search.

% knowledge-based

\noindent \textbf{Heuristic Approaches:}
%By imitating traditional manual feature engineering, the work presented in \citep{cheng2011automated} extracts features from semantic knowledge bases. However, it relies heavily on domain knowledge.
% expansion-reduction
Deep Feature Synthesis (DFS), the component of Data Science Machine \citep{kanter2015deep}, first enumerates all transformations on all features and then performs feature selection directly based on the improvement of model performance.
One Button Machine \citep{lam2017one} adopts a similar approach.
However, this \textit{expansion-reduction} approach suffers from a severe computational performance bottleneck due to the huge feature evaluation overhead.
% performance-guided
To avoid enumerating the entire feature space, Cognito \citep{khurana2016automating} introduces a tree-like exploration of feature space and presents handcrafted heuristics traversal strategies such as breadth-first search and depth-first search.
AutoFeat \citep{horn2019autofeat} iteratively subsamples features using beam search.
However, heuristic approaches cannot learn from past experiences and thus has a low search efficiency.

% learning-based
\noindent \textbf{Learning-Based Approaches:} To explore feature space efficiently, learning-based AutoFE methods have been proposed.
% Recently, the learning-based AutoFE methods have been proposed to efficiently explore the feature space.
LFE \citep{nargesian2017learning} trains an MLP and recommends the most likely useful transformation for each raw feature. However, it does not support transformation composition and works only for classification tasks.
% transformation graph
TransGraph \citep{khurana2018feature} trains a Q-learning agent to decide which transformation should be applied.
%%
%TransGraph also suffers from the feature explosion problem.
%
Due to performing each transformation on all features, TransGraph suffers from feature explosion and low computational efficiency.
% Since the number of new features to be evaluated each time is linearly correlated with the number of raw features, it still suffers from the feature explosion problem.

% 介绍NAS
% 例句: Lately, automated machine learning (AutoML) [22] has aroused great research interests from both academia and industry
\noindent \textbf{NAS-Based Approaches:}  Neural Architecture Search \citep{elsken2019neural} has aroused significant research interests in the field of AutoML \citep{he2020automl}.
The reinforcement learning-based NAS method \citep{zoph2016neural} views the structure of a neural network as a variable-length string.
Then, it uses a recurrent network as the controller to generate such strings and trains the controller with policy gradient.
% 结合NAS思路的AutoFE算法
This approach can be adopted into AutoFE.
For instance, NFS \citep{chen2019neural}, the current SOTA AutoFE method, utilizes several RNN-based controllers to generate transformation sequences for each raw feature.
%To train the RNN controllers NFS views the evaluation results of all sub-sequences viewed as the rewards.
However, evaluating enormous sequences results in substantial computational overhead. 
% The evaluation results of all sub-sequences are viewed as the rewards for training the RNN controllers based on policy gradient \citep{sutton2000policy}.
% A large number of sequences to be evaluated results in huge computational overhead.
% However, all sub-sequences of the feature transformation sequences need to be evaluated, resulting in huge computational overhead.
Most importantly, due to the side effects of reducing binary transformations to unary ones, NFS cannot generate complex features like $\frac{A + B}{C - D}$.
% as NAS is modeled as a Markov Decision Process, credits are assigned to structural decisions with temporal-difference (TD) learning (Sutton et al., 1998), whose efficiency and interpretability suffer from delayed rewards

% Differentiable NAS
To improve the computational efficiency of NAS, differentiable methods have been proposed.
DARTS \citep{darts} relaxes the categorical choice to a softmax over all possible operations, leading to a differentiable learning objective.
NAO \citep{nao} maps the discrete architecture space to a continuous hidden space and optimizes existing architectures in the continuous space.

The differentiable NAS methods bring more inspiration to AutoFE.
%AutoFIS \citep{liu2020autofis} employs DARTS to solve the problem of automated feature interaction selection in the FM algorithm \citep{rendle2010factorization}.
% AutoFIS \citep{liu2020autofis}, working for feature interaction selection in FM \citep{rendle2010factorization}, is a related work of DARTS in AutoFE.
In this paper, we propose the first differentiable AutoFE method called \nfo{}, which can efficiently construct useful low-order and high-order features with much fewer feature evaluations.

\section{Methodology}

% 形式化Feature Generate
\subsection{Problem Formulation}

Let $D = \langle F, y \rangle$ be a dataset with a target vector $y$ and $n$ $d$-dimensional instances $F=\lbrace f_1, \cdots, f_d \rbrace$, where $f_i \in \mathcal{R}^n$ is the $i$-th raw feature.
We denote the performance of the machine learning model $M$ that is learned from $D$ and measured by an evaluation metric $L$ (e.g., F1-score or mean squared error) as $L_M \left( F,y\right)$.
Without loss of generality, the higher $L_M$ indicates better model performance.

Furthermore, we apply the composition of transformations $t \in \mathcal{R}^{n} \times \cdots \times \mathcal{R}^{n} \rightarrow \mathcal{R}^{n}$ to features for constructing new features.
% Let $o$ denote the arity of the transformation $t$, we construct a new feature $
% f_i = t\left( f_{i_1}, \cdots, f_{i_o} \right)$, where $f_{i_j}$ denotes the $j$-th input of $t$ to construct $f_i$ for $j \in  \{1, \cdots, o\}$.
% For example, we use the composition of the unary transformation $\textit{square}$ and the binary transformation $\textit{divide}$ to construct \textit{BMI} (Body Mass Index) by $\textit{divide} \left( \textit{weight}, \textit{square} \left( \textit{height} \right) \right)$ with the raw features \textit{weight} and \textit{height}.
% modified by zgh 0420
Let $o$ denote the arity of the transformation $t$, we construct a new feature $ \hat{f} = t\left(\hat{f}_{1}, \cdots,\hat{f}_{o} \right)$, where $\hat{f}_{j}$ denotes the $j$-th input of $t$ to construct $\hat{f}$  for $j \in  \{1, \cdots, o\}$.
Given a set of transformations with different arities $T = \lbrace t_{1}, \cdots, t_{m} \rbrace$, we define the feature space $F^T$ as follows: $\forall \hat{f} \in F^T$, $\hat{f}$ satisfies any of the following conditions:
\begin{itemize}
\item $\hat{f} \ \in \ F$
\item $\exists t \in T, \hat{f} = t \left(\hat{f}_{1}, \cdots, \hat{f}_{o} \right)$, where $\hat{f}_{1}, \cdots, \hat{f}_{o} \in F^T$
\end{itemize}

% We use a tree-like structure to represent extracted features universally.

% % 通过闭包定义法来定义特征空间
% Given a set of transformations with different arities $T = \lbrace t_{1}, \cdots, t_{m} \rbrace$, we define the feature space $F^T$ as follows: $\forall f_i \in F^T$, $f_i$ satisfies any of the following conditions:

% zgh 0420通过闭包定义法来定义特征空间

% we use $o$ to describe the number of features $t$ takes as input and use $f_{i_j}$ to represent the $j$-th input feature of $o$-ary $t$ used to construct $f_i$ . Then, we define that a feature $f_i$ belongs to feature space $F^{T}$, which is generated by $T$ from $F$, if any of the following conditions is met:

% \begin{itemize}
% \item $f_i \ \in \ F$
% \item $\exists t \in T,\ f_i = t \left( f_{i_1}, \cdots, f_{i_o} \right)$, where $f_{i_1}, \cdots, f_{i_o} \in F^T$
% \end{itemize}

%zgh 0420

Formally, let $\alpha(\hat{f})$ denote the $\textit{order}$ of the feature $\hat{f} \in F^T$, $\alpha(\hat{f})$ can be defined as:
\begin{equation}
\alpha (\hat{f})=
\begin{cases}
1 + \mathop{\max}_j \alpha \left( \hat{f}_{j} \right) & \hat{f} = t(\hat{f}_{1}, \cdots, \hat{f}_{o})\\
0& \hat{f} \in F
\end{cases}
\end{equation}
For example, we use the composition of the unary transformation $\textit{square}$ and the binary transformation $\textit{divide}$ to construct \textit{BMI} (Body Mass Index), whose order is 2, by $\textit{divide} \left( \textit{weight}, \textit{square} \left( \textit{height} \right) \right)$ with the raw features \textit{weight} and \textit{height}.

Therefore, the goal of AutoFE is to find the set of constructed features $F^{*}$ that can achieve the best performance:
\begin{equation}
    F^{*} = \mathop{\arg\max}_{\hat{F}} \ L_M (F \cup \hat{F}, y), \text{ s.t. } \hat{F} \subset F^T
    \label{eq:opt_original}
\end{equation}

In practice, we limit the order of features and search in the feature space $F_k^T = \lbrace \hat{f} \ | \ \hat{f} \in F^T \land \alpha(\hat{f}) \leq k \rbrace$ since the size of the original space is infinite (i.e., $\lvert F^T \rvert = \aleph_0 $).
we explore $F_k^T$ and search for top features ranked by the performance metric $L_M \left( F \cup \lbrace \hat{f} \rbrace, y\right)$ as $F^{*}$.
Moreover, similar to most existing AutoFE methods (e.g., NFS \citep{chen2019neural},  \citep{nargesian2017learning}, and TransGraph~\citep{khurana2018feature}), we also append the constructed features to $F$ to maximize the modeling performance for a given algorithm.

%To ensure the accuracy of feature performance, \nfo{} returns the final features set containing all original features $F$ as $F^{*}$.

% To eliminate the combinatorial explosion problem, we further prune the feature space by preserving only top features in $F_k^T$ and use a greedy algorithm to search for the optimal combination in the pruned space.
% Thus, the problem can be decomposed into two steps:
% \begin{enumerate}
% \item Explore the entire feature space $F_k^T$, and search for top features $F_{\text{top}} \subset F_k^T$ ranked by the performance $L_M \left( F \cup \lbrace f \rbrace, y\right)$, where $f \in F_k^T$.
% \item Exploit the pruned space $F_{\text{top}}$ to find the optimal combination of features $F^{*} = \mathop{\arg\min}_{\hat{F}} \ L_M (F \cup \hat{F}, y), \text{ s.t. } \hat{F} \subseteq F_{\text{top}}$.
% % \item Randomly sample combinations in the pruned space $F_{\text{top}}$ to find the best set of features $F^{*} = \mathop{\arg\min}_{\hat{F}} \ L_M (F \cup \hat{F}, y), \text{ s.t. } \hat{F} \subseteq F_{\text{top}}$.
% \end{enumerate}

% 结合EA算法, 整体框架
\subsection{Overview of DIFER}
\begin{figure}[ht]
        \centering
        \includegraphics[width=1.0\textwidth]{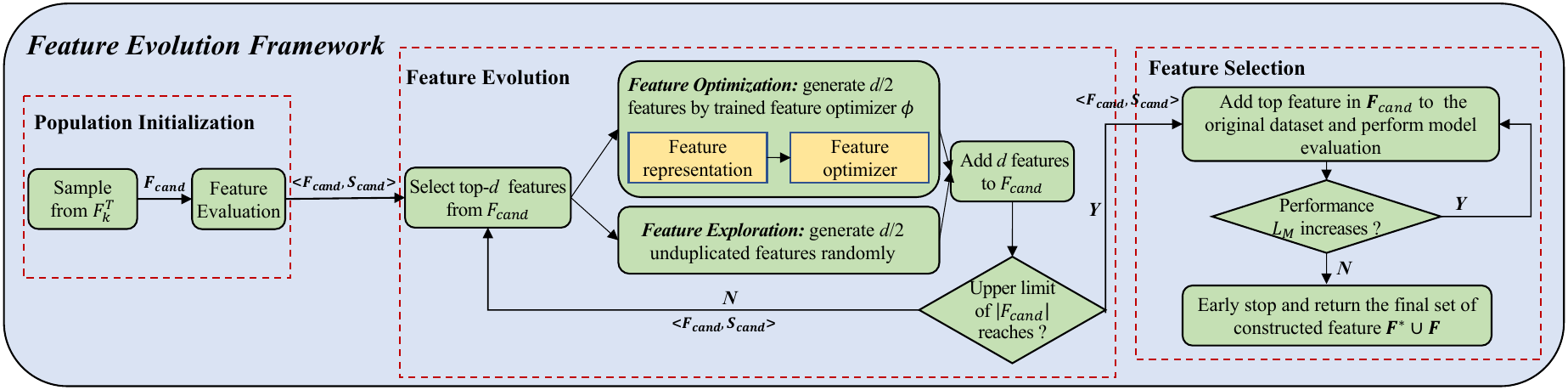}
        \vspace{-2ex}
        \caption{Overview of \nfo{}.}
        \label{fig:workflow}
    \vspace{-3ex}
\end{figure}

As shown in Figure~\ref{fig:workflow}, we propose an evolutionary framework to achieve AutoFE.
The overall framework is divided into three phases: population initialization, feature evolution, and feature selection.

The population initialization phase constructs feature set $F_{\text{cand}}$ by randomly sampling features from $F_k^T$.
We train a machine learner $M$, which takes instances as input and predicts the labels $y$, from scratch and evaluate its performance as the performance score of the feature $ L_M(F \cup \lbrace \hat{f} \rbrace, y)$.
Then, we can get the score set $S_{\text{cand}} = \lbrace L_M(F \cup \lbrace \hat{f} \rbrace, y) | \ \hat{f} \in F_{\text{cand}} \rbrace$.

The feature evolution phase aims to construct new features iteratively. 
In each iteration, we first select top-$d$ features from $F_{\text{cand}}$ according to $S_{\text{cand}}$. 
To enhance the diversity of evolution, we take two different approaches to generate new features at the same time.
One way is to perform gradient-based optimization based on the feature optimizer and add $d/2$ optimized features to $F_{\text{cand}}$ (i.e., exploitation). 
The other way is to add $d/2$ unduplicated randomly-generated features to $F_{\text{cand}}$ for exploration.
The process of feature evolution is repeated until a maximum number of feature evaluations is reached. 
In the feature optimization process, the two key components are the parse-tree-based feature representation and the gradient-directed feature optimizer that consists of an encoder, a predictor, and a decoder.
% Figure~\ref{fig:optimization} illustrates the gradient-directed feature optimizer, which consists of an encoder, a predictor, and a decoder.
%
Due to its flexibility in the optimization of complex feature transformation, the encoder-predictor-decoder-based feature optimizer is suitable for the AutoFE problem.

After the feature evolution phase, we select top features from $F_{\text{cand}}$ and add them to the original dataset. 
The number of added features is adaptively determined with an early-stopping mechanism. 
When the model performance no longer increases, we stop adding features to the original dataset. 
% When the model performance no longer increases with a specific patience (i.e., 3), we stop adding features to the original dataset. 
% %
% Moreover, the maximum value of $k$ is $\min$(d, the total number of original features).

% $F_{\text{cand}}$ and  $S_{\text{cand}}$ are fed into the next step for the training of feature optimizer.

% The encoder-predictor-decoder-based optimizer is suitable for the AutoFE problem due to its flexibility in the optimization of complex feature transformation.
%Next, we introduce the two key components.

\paragraph{Case Study}
we show the process of \nfo{} using the dataset \textit{PimaIndian} as an example.
\nfo{} first initializes the population $\langle F_{cand}, S_{cand} \rangle$ by random sampling and evaluating features from $F_k^T$.
Then, the feature optimizer is trained on the population. 
The detailed training process of feature optimizer is introduced in Section~\ref{ss:feature_optimizer}.

%
% As shown in Figure~\ref{fig:workflow}, the next phase is the feature evolution that consists of feature optimization and feature exploration.
In the feature optimization process, taking the feature $\frac{\textit{min\_max}(\textit{BloodPressure}) }{\textit{Insulin}}$ as an example, we introduce how the input feature is optimized to get a better feature.
As mentioned in Section~\ref{s:feat-repr}, the feature is first parsed as a tree and traversed to the string <Insulin,\textit{Reciprocal},BloodPressure,\textit{MinMax},\textit{Multiply}>.
The feature optimizer $\psi$ maps it into the continuous vector space as $e_x$ via the encoder $\psi_e$, optimizes the embedding $e_x$ along the gradient direction induced by the predictor $\psi_p$. The string <Insulin, Pregnancies, \textit{AbsRoot}, \textit{Multiply}, \textit{Reciprocal},BloodPressure,\textit{MinMax},\textit{Multiply}> is recovered from the optimized embedding $e_{x^{'}}$ by the decoder $\psi_d$.
The recovered string is translated to $\frac{\textit{min\_max}(\textit{BloodPressure}) }{\sqrt{\lvert \textit{Pregnancies} \rvert} \cdot \textit{Insulin}}$.
%with no ambiguity via the translation process in Appendix A.
% The feature evolution is repeated until the number of feature evaluations reaches the upper limit, and then the final features $F^{*}$ is selected by \nfo{} from $F_{cand}$.
% Finally, we improve the performance $L_M(F \cup F^{*}, y)$ by constructing features $F^{*}$ on \textit{PimaIndian}.

\begin{figure}[t]
    \centering
    \includegraphics[width=1.0\linewidth]{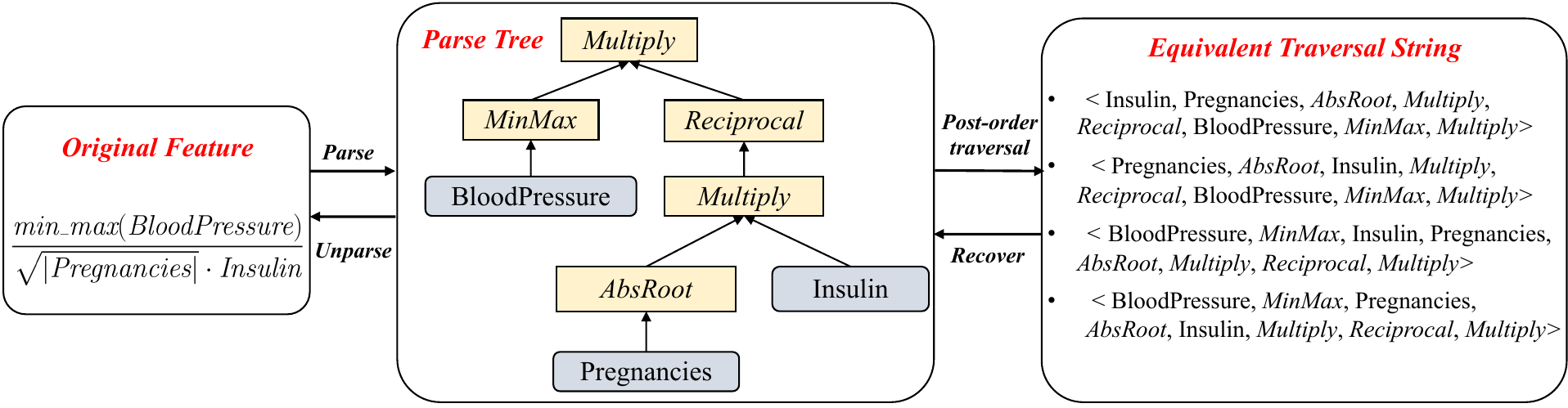}
    \caption
    {Parse tree and post-order traversal strings of the feature
    $\frac{\textit{min\_max}(\textit{BloodPressure}) }{\sqrt{\lvert \textit{Pregnancies} \rvert} \cdot \textit{Insulin}}$
    in \textit{PimaIndian}.}
    \label{fig:tree}
    \vspace{-1ex}
\end{figure}

\subsection{Feature Representation}
\label{s:feat-repr}

As shown in Figure~\ref{fig:tree}, we design a tree-like structure called \textit{parse tree} to represent constructed features.
Compared with the sequence representation in NFS~\citep{chen2019neural} and NAO~\citep{nao}, the parse tree is more flexible and expressive, which can represent complex $n$-ary feature transformation operation like $\frac{A + B}{C - D}$.
The internal node in the parse tree indicates the transformation and the leaf node indicates the raw feature.
We employ reversible post-order traversal to convert the parse tree into equivalent traversal string $x$ as input to the encoder.
The traversal string in Figure~\ref{fig:tree} shows an example where each word-based token (i.e., the original feature and the transformation) is separated by a comma.
Let $x_r$ denote each token in the traversal string, where $r \in \{1 \cdots \lvert x \rvert \}$.
Note that the relationship between the parse tree and the traversal string is one-to-many.
When there are transformations where the input order is meaningless (e.g. $\textit{mul}(a, b) == \textit{mul}(b, a)$), the same parse tree can be converted into multiple equivalent strings.
This nature can be viewed as a way of data augmentation when training the feature optimizer.
Due to the fixed arity of each transformation, the optimized traversal string can be recovered to a parse tree with no ambiguity.
The translation process can be found in Appendix~\ref{s:bnf}.
\subsection{Feature Optimizer}
\label{ss:feature_optimizer}

\begin{figure}
    \begin{subfigure}[ht]{\linewidth}
        \centering
        \includegraphics[width=0.7\textwidth]{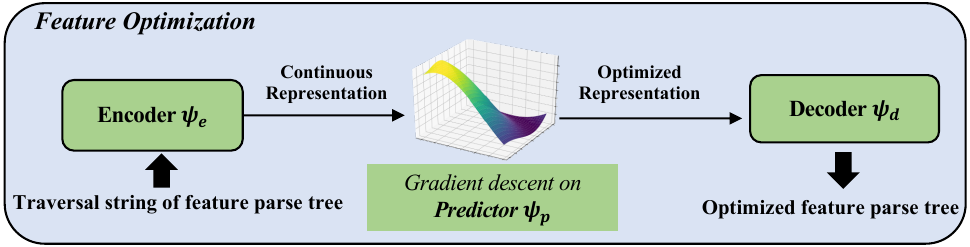}
        \caption{Feature optimization.}
        \label{fig:optimization}
    \end{subfigure}
    \hfill
    \begin{subfigure}[ht]{\linewidth}
        \centering
        \includegraphics[width=0.7\textwidth]{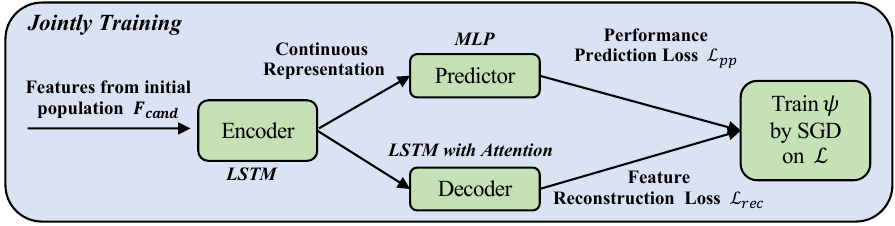}
        \caption{Jointly-training of feature optimizer.}
        \label{fig:co-training}
    \end{subfigure}
    \caption{Feature Optimizer of \nfo{}.}
    \vspace{-3ex}
\end{figure}

\nfo{} employs a feature optimizer to construct new features based on the existing features.
The feature optimization process is shown in Figure~\ref{fig:optimization}.
Specifically, the feature optimizer $\psi$ consists of an encoder $\psi_e$, a performance predictor $\psi_p$, and a decoder $\psi_d$.
After jointly-training the feature optimizer for convergence, $\psi$ maps features into the continuous vector space via $\psi_e$, optimizes the embedding along the gradient direction induced by $\psi_p$, and recovers better features from the optimized embedding by $\psi_d$.

% \begin{figure}[t]
% \centering
% \includegraphics[width=0.49\textwidth]{figures/tree.pdf}
% \caption
% {Parse tree and post-order traversal strings of the feature
% $\frac{\textit{min\_max}(\textit{BloodPressure}) }{\sqrt{\textit{Pregnancies}^2} \cdot \textit{Insulin}}$
% in the \textit{PimaIndian} dataset.}
% \label{fig:tree}
% \end{figure}

\paragraph{Encoder} The encoder $\psi_e$ maps the post-order traversal string $x \in \mathcal{X}$ to a continuous embedding $e_x \in \mathcal{E} \subset \mathcal{R}^{emb\_dim}$.
Since the traversal string $x$ is a variable-length sequence, we use LSTM (Long Short-Term Memory) \citep{hochreiter1997long} as the encoder.
% which can take variable-length sequences as input.
By the sum-pooling technique, the sum of all hidden states $H_{x}=\{h_1, h_2, \cdots, h_{\lvert x \rvert}\}$ of the LSTM as the feature's continuous representation $e_x$.
\iffalse
Inspired by the message aggregation function in GNNs (Graph Neural Network) \citep{wu2020comprehensivegraph}, we leverage the sum-pooling technique and view the sum $e_x$ of all hidden states $H_{x}=\{h_1, h_2, \cdots, h_{\lvert x \rvert}\}$ of the LSTM as the continuous representation of $x$.
Compared to the average of all hidden states used in NAO \citep{nao}, $e_x$ can contain the information about the length of sequences just like the \textit{sum} message aggregation function.
\fi
\paragraph{Predictor} The predictor $\psi_p \in \mathcal{E} \rightarrow \mathcal{R}$ maps the continuous representation $e_x$ into its score $s_x$ measured by $L_M (F \cup \lbrace \hat{f}_x \rbrace, y)$.
We employ a 5-layer fully-connected MLP as $\psi_p$.
\paragraph{Decoder} The decoder $\psi_d$ maps the embedding to the discrete feature space, i.e., the post-order traversal string of the optimized feature.
According to the classical sequence-to-sequence method, we employ an LSTM with the attention mechanism \citep{bahdanau2014neural} as the decoder $\psi_d \in \mathcal{E} \rightarrow \mathcal{X}$, which takes $e_x$ as the initial hidden state and all hidden states $H_x$ in the encoder as the input of each timestamp.
%According to the many-to-one relationship between the traversal string and the parse tree, we can recover the parse tree from the traversal string with no ambiguity.

% \subsection{Co-Training of Feature Optimizer}
\paragraph{Jointly-Training} To train the optimizer efficiently, we propose a jointly-training method based on a joint loss.
%A co-training strategy is proposed to train the whole feature optimizer.
%
The training dataset is the initial evaluated population $\langle F_{\text{cand}}, S_{\text{cand}} \rangle$.
As shown in Figure~\ref{fig:co-training}, we design a joint loss function that takes both the performance prediction loss $\mathcal{L}_{pp}$ and the structure reconstruction loss $\mathcal{L}_{rec}$ into account.
The value of the hyperparameter $\lambda$ that balances $\mathcal{L}_{pp}$ and $\mathcal{L}_{rec}$ is determined adaptively (see Appendix~\ref{s:adaptive-loss}).
%by relative loss weight $\lambda$:
% We take into account both the performance prediction loss $\mathcal{L}_{pp}$ and the structure reconstruction loss $\mathcal{L}_{rec}$ using a unified loss function.

\vspace{-3ex}
\begin{equation}
\mathcal{L} = \lambda \mathcal{L}_{pp} + \mathcal{L}_{rec}
\text{, where }\mathcal{L}_{pp} = \sum_x \left( s_x - \psi_p \left( \psi_e \left(  x\right) \right) \right)^2 \text{ and }
\mathcal{L}_{rec} = -\sum_x \sum_{r=1}^{\lvert x \rvert} \log{ P_{\psi_{d}}(x_r|\psi_e(x)) }
\label{eq:loss}
\end{equation}

% evolution 获取新特征, 谈到相似度度量空间, 泰勒展开
\subsection{Feature Optimization}

% \begin{figure}[t]
% \centering
% \includegraphics[width=0.40\textwidth]{figures/emb-space.pdf}
% \caption{Optimize feature in vector space by gradient descent}
% \label{fig:emb-space}
% \end{figure}

% feature string & data augmentation
% As mentioned in Figure \ref{fig:tree}, the feature $f \in F^T$ is a tree structure with the transformations as internal nodes and the raw features as leaf nodes. We use the post-order traversal strings of trees to specify the features. The string for each tree is unique because the transformations take a fixed number of features as input.
% Furthermore, the order of inputs for some transformations are meaningless such as $\textit{add}(x, y) == \textit{add}(y, x)$.
% According to this nature, the same feature transformation tree can be converted into multiple strings as a way of data augmentation.

%With SGD, the training proceeds in steps, and at each step we consider a mini-batch x1...m of size m. The mini-batch is used to approximate the gradient of the loss function concerning the parameters, by computing
After the convergence of the feature optimizer, we directly optimize the feature embedding $e_x$ in the continuous space by performing gradient ascent and then decode the optimized embedding into a new feature $x^{'}$.
% One way to find the global optimal solution is to gradually approach the goal through iteration, like gradient ascent. Similarly, we exploit searched features to get top features in the feature space.

Starting from the constructed feature $x$, we optimize its embedding $e_x$ to get a better embedding along the gradient direction induced by the predictor $\psi_p$:
\begin{gather}
    e_{x^{'}} = \sum_{h_r \in H_x} \left( h_r + \eta \frac{\partial \psi_p}{\partial h_r} \right)
    % h_r^{'} = h_r - \eta \frac{\partial f_p}{\partial h_r}, \ \forall h_r \in H_x
    \label{eq:evolution}
\end{gather}

However, due to the nature that the corresponding parse tree of a feature $\hat{f}_x$ may have several equivalent post-order traversal strings $X = \lbrace x^{(1)}, x^{(2)}, \cdots x^{(n)} \rbrace$, the strings in $X$ are highly similar in the continuous space.
% The neighborhood of $X$ in the continuous space can be represented as follows:
% \begin{equation}
%     \exists \epsilon,\ \forall x^{(i)} \in X, \Vert e_{x^{(i)}} - \frac{1}{n} \sum_{x^{(j)} \in X} e_{x^{(j)}} \Vert_2 \leq \epsilon
%     \label{eq:neighbour}
% \end{equation}
% , where $\epsilon$ is just a variable used to inscribe the property, not a hyperparameter.
After one step of gradient ascent, the decoded string of $e_{x^{'}}$ may still be in $X$.
% As shown in Figure \ref{fig:emb-space}, $e_{x^{'}}$ may still be in the neighborhood of $X$ after gradient ascent.
Thus, we may get the same parse tree.
We call $\eta$ in Equation~\eqref{eq:evolution} the evolution rate.
Increasing the evolution rate $\eta$ can solve this problem to some extent~\citep{nao}.
% Since the learning rate $\eta$ makes $\Delta x$ too large which violates the objective of gradient ascent, NAO cannot guarantee that $\psi_p (x + \Delta x) < \psi_p (x) $ and finally gets a worse embedding in the continuous space.
However, a large evolution rate would violate the preconditions of gradient ascent, resulting in no guarantee that $\psi_p (x + \Delta x) > \psi_p (x) $.
% In fact, when performing gradient ascent on the objective function $\psi_p$, the first-order Taylor polynomial is used:
% % However, the core of the gradient ascent method to ensure that the value of the objective function $\psi_p$ decreases is an approximation based on first-order Taylor formula:
% \begin{equation}
%     \psi_p (x + \Delta x) = \psi_p (x) + \langle \nabla \psi_p(x), \Delta x \rangle + o(\Delta x)
%     \label{taylor-expansion}
% \end{equation}
% where $\Delta x = -\eta \cdot \nabla \psi_p (x)$.
% %$o(\Delta x)$ denotes a quantity decaying faster than $\Delta$ toward 0.
% The core to ensure that $\psi_p$ decreases is the use of a small enough learning rate $\eta$ to satisfy $\lim_{\Delta x \to 0} o(\Delta x) = 0$, so that
% \begin{equation}
%     \psi_p (x + \Delta x) = \psi_p(x) - \eta \langle \nabla \psi_p(x),\nabla \psi_p(x) \rangle  < \psi_p(x)
%     \label{taylor}
% \end{equation}

% However, when performing gradient ascent on the objective function $\psi_p$, the learning rate need to be small enough to make sure that $\lim_{\Delta x \to 0} o(\Delta x) = 0$.
% Otherwise, the first-order Taylor approximation in Equation \eqref{taylor} does not hold:
% \begin{equation}
%     \psi_p(x + \Delta x) = \psi_p(x) - \psi_p^{'}(x) \cdot \eta \psi_p^{'}\left( x \right) + o(\Delta x) < \psi_p(x)
%     \label{taylor}
% \end{equation}

\paragraph{Multi-step gradient ascent} To address this problem, we propose a straightforward but effective strategy.
Specifically, we apply the optimization process in Equation \eqref{eq:evolution} multiple times with a small evolution rate $\eta$ until we get new parse trees.
As a result, the number of times the optimization process (i.e., steps of gradient ascent) is adaptively determined.
We refer to the overall process as feature optimization.

\section{Experiments}

% In this section, we conduct extensive experiments on public datasets to answer the following questions:

% \begin{itemize}
% \item \textbf{RQ1}: How effective is the proposed \nfo{} approach?
% \item \textbf{RQ2}: How efficient is \nfo{} to explore the feature space?
% \item \textbf{RQ3}: How effective are the high-order features constructed by \nfo{}?
% \item \textbf{RQ4}: Can \nfo{} improve the performance of different machine learning algorithms?

% %\item \textbf{RQ5}: Is \nfo{} robust to hyperparameters?
% \end{itemize}

\subsection{Experimental Setting}
\label{sec:experiment-setting}

As with the SOTA method NFS~\citep{chen2019neural}, we use 25 public datasets from OpenML~\citep{OpenML}, UCI repository~\citep{UCI}, and \citet{Kaggle}. 
There are 15 classification (C) datasets and 10 regression (R) datasets that have various numbers of features (5 to 10936) and instances (100 to 30000).
In all experiments, we set the max order $k$ to 5 except in RQ3 and utilize 9 transformation functions totally.
Moreover, to ensure the fairness, all methods except LFE~\citep{nargesian2017learning} have the same feature transformation space.
\begin{itemize}[topsep=0.3ex,parsep=0ex]
\item Unary transformation: \textit{logarithm}, \textit{square root}, \textit{min-max normalization}, and \textit{reciprocal}
\item Binary transformation: \textit{addition}, \textit{subtraction}, \textit{multiplication}, \textit{division}, and \textit{modulo}
\end{itemize}

% employ the widely used technique of cross-validation to estimate it
All experiments are run using Tesla K80 (GPU) and Intel(R) Xeon(R) CPU E5-2630 v2 instances.
To evaluate the AutoFE method, we use the performance metric $\left(1 - \left( \textit{relative absolute error} \right)\right)$ \citep{shcherbakov2013survey} for the regression task and \textit{f1-score} for the classification task.
5-fold cross validation using random stratified sampling is employed and the average result is reported.
Except that different ML algorithms are used in RQ4, we utilize Random Forest as default.
We use scikit-learn as the machine learning algorithm library and employ PyTorch to implement the feature optimizer, including LSTM-based encoder and decoder, MLP-based predictor.

%The hyperparameters can be divided according to three steps.
In the initialization step of \nfo{}, we randomly select 512 features as the initial population.
Both the encoder and decoder of the feature optimizer are implemented as a one-layer LSTM.
%
% Moreover, Dot-Product attention is used in the decoder.
We empirically set the embedding size of each token in the traversal string and the size of the hidden state to 512.
The predictor is a 5-layer MLP where the number of hidden units in each layer is 1024.
To train the feature optimizer, we choose the Adam optimizer~\citep{kingma2014adam} with a learning rate of 0.001 and a weight decay of 0.0001. 
The number of epochs is 400, and the batch size is 128.
%, and the dropout rate is 0.5.
%
Early stopping is employed with a patience of 10.

In each feature evolution iteration, the value of $d$ is empirically set to be the minimum between top 20\% of the initial population size and the total number of original features.
The feature evolution runs until the number of feature evaluations reaches the upper limit of 4096. 
When optimizing the feature embedding in Equation \eqref{eq:evolution}, we perform gradient ascent with an evolution rate $\eta$ of 0.0001.
Moreover, we use the same hyperparameters for all datasets.
The robustness experiments with different hyperparameters can be found in Appendix~\ref{s:hyper}.

\begin{table*}[t!]
	\centering
    \caption{Comparison between \nfo{} and the existing AutoFE methods (The datasets are sorted based on the evaluation time. $^{\dag}$ the results obtained using the open-sourced code, $*$ denotes statistically significant improvement measured by Friedman test and Nemenyi post-hoc test with $p$-value $<0.05$. $\mathcal{T}$ indicates the total runtime. Inst. is short for Instance, Feat. is short for Feature, \textit{Err.} indicates failure due to out of memory when running the open-source code).}

	\scalebox{0.76}{
		\begin{tabular}{ c | c c | r r r r r r | r r}
			\toprule
			Dataset&   C/R&    Inst.$\backslash$Feat.&   Raw&   Random&  DFS$^{\dag}$ & AutoFeat$^{\dag}$ & NFS$^{\dag}$&    \nfo{}$^{*}$ &  $\mathcal{T}_{NFS}$  & $\mathcal{T}_{\nfo{}}$ \\
			\midrule

            Housing Boston&    R&  506$\backslash$13& 0.4336&  0.4446&  0.3412& 0.4688& \textbf{0.5013}&   0.4944 & \textbf{566.42} & 982.15 \\

			Bikeshare DC&  R&  10886$\backslash$11& 0.8200&  0.8436& 0.8214& 0.8498& 0.9746&  \textbf{0.9813} & \textbf{595.57} &  1040.96 \\

			Airfoil&  R&  1503$\backslash$5&  0.4962&  0.5733& 0.4346&  0.5955& 0.6163&  \textbf{0.6242} & \textbf{603.80} & 1066.93 \\
			
			Openml\_586&    R&  1000$\backslash$25&  0.6617&  0.6511& 0.6501& 0.7278& 0.7401& \textbf{0.7683} & 1722.49 & \textbf{1013.57}  \\
			
			Openml\_589&    R&  1000$\backslash$25&  0.6484&  0.6422& 0.6356& 0.6864& 0.7141&  \textbf{0.7727} & 1726.04 & \textbf{1005.18}  \\

			Openml\_637&     R&  1000$\backslash$25& 0.5136&  0.5268& 0.5191 & 0.5763& 0.5693&  \textbf{0.6343} & 1411.79 & \textbf{1028.14} \\

			Openml\_618&    R&  1000$\backslash$50&  0.6267&  0.6167& 0.6343& 0.6324& 0.6400&  \textbf{0.6603} & 3159.47 & \textbf{1020.72} \\

			Openml\_607&    R&  1000$\backslash$50&  0.6344&  0.6285& 0.6388& 0.6699& 0.6870&  \textbf{0.6918} & 2990.91 & \textbf{1032.40} \\
			
			Openml\_616&     R&  500$\backslash$ 50&  0.5747&  0.5714& 0.5717& 0.6027& 0.5915&  \textbf{0.6554}& 1511.58 & \textbf{1030.57} \\
			
			Openml\_620&    R&  1000$\backslash$25& 0.6336&  0.6178&  0.6263& 0.6874& 0.6749&  \textbf{0.7442} & 1686.78 & \textbf{1047.37} \\
			
			\midrule
			
			Hepatitis&      C&  155$\backslash$6& 0.7860&  0.8300& 0.8258& 0.7677& 0.8774&  \textbf{0.8839} & \textbf{355.76} & 1045.77 \\
			
			Fertility&     C&  100$\backslash$9& 0.8530&  0.8300& 0.7500& 0.7900&  0.8700& \textbf{0.9098} & \textbf{362.38} & 1054.51 \\

			SpectF&    C&  267$\backslash$44&  0.7750&  0.8277& 0.7906& 0.8161& 0.8501&  \textbf{0.8612} & \textbf{386.39} & 933.45 \\

			Megawatt1& C&  253$\backslash$37& 0.8890&  0.8973& 0.8773& 0.8893&  0.9130&  \textbf{0.9171} & \textbf{404.33}  & 1024.95\\

			Ionosphere&     C&  351$\backslash$34& 0.9233&  0.9344&  0.9175& 0.9117& 0.9516& \textbf{0.9770} & \textbf{421.50} & 1036.01 \\

			German Credit&     C&  1001$\backslash$24& 0.7410& 0.7550& 0.7490& 0.7600 & \textbf{0.7818}& 0.7770 & \textbf{433.39} & 1043.06 \\

			Credit-a&    C&  690$\backslash$6& 0.8377&  0.8449& 0.8188& 0.8391& 0.8652&  \textbf{0.8826} & \textbf{435.14} & 992.91 \\

			PimaIndian&  C&  768$\backslash$8& 0.7566&  0.7566& 0.7501& 0.7631& 0.7839&  \textbf{0.7865} & \textbf{435.10} & 1007.30\\

			Messidor\_features&    C&  1150$\backslash$19& 0.6584&  0.6878& 0.6724& 0.7359&  0.7461&  \textbf{0.7576} & \textbf{555.62} & 1069.04 \\

			Wine Quality Red&  C&  999$\backslash$12& 0.5317&  0.5641& 0.5478& 0.5241& \textbf{0.5841}&  0.5824 & \textbf{587.77} & 1033.29  \\
			
			Wine Quality White&   C&  4900$\backslash$12& 0.4941&  0.4930& 0.4882& 0.5023& 0.5150& \textbf{0.5155} & 1278.61 & \textbf{1016.35} \\
			
			SpamBase&   C&  4601$\backslash$57& 0.9102&  0.9237&  0.9102& 0.9237& 0.9296&  \textbf{0.9339} & 993.92& \textbf{959.03} \\
			
			AP-omentum-ovary & C& 275$\backslash$10936 & 0.7636 & 0.7100 & 0.7250 & \textit{Err.} & 0.8640 & \textbf{0.8726} & 4183.75 & \textbf{1441.01} \\
			
			Credit Default&    C&  30000$\backslash$25& 0.8037&  0.8060&  0.8059& 0.8060 & 0.8049&  \textbf{0.8096} & 9253.70 & \textbf{1204.99} \\

			gisette & C& 2100$\backslash$5000 & 0.9261 & 0.8710 & 0.7410 & \textit{Err.} & 0.9590 & \textbf{0.9635} & 18877.07 & \textbf{1646.19} \\
			
			\midrule
			Upper Limit of Eval. Num. & & & & & & & 160,000& \textbf{4,096} &  &\\
			\bottomrule
		\end{tabular}

	}
	 \label{tab:comp1}
	 \vspace{-3ex}
\end{table*}

\subsection{Effectiveness of \nfo{} (RQ1)}

In this subsection, we demonstrate the effectiveness of \nfo{}.
We compare \nfo{} on 25 datasets with the SOTA and baseline methods, including: (a) Raw: raw dataset without any transformation; (b) Random: randomly applying transformations to each raw feature; (c) DFS~\citep{kanter2015deep}: a well-known \textit{expansion-reduction} method; (d) AutoFeat~\citep{horn2019autofeat}: a popular Python library for automated feature engineering and selection; (e) LFE~\citep{nargesian2017learning}: recommend the most promising transformation for each feature using MLP; (f) NFS~\citep{chen2019neural}: the SOTA AutoFE method that achieves better performance than other existing approaches (e.g., \citet{khurana2018feature}).
The experimental settings of these methods, such as the set of transformations, the max feature order, and the evaluation metrics are the same as \nfo{}.

Table \ref{tab:comp1} shows the comparison results between \nfo{} and the existing methods. Moreover, since LEF can only deal with the classification task and the source code is not available, we directly use the best results reported in the original paper~\citep{nargesian2017learning}. The comparison results between \nfo{} and LEF are shown in Table 2. From Table~\ref{tab:comp1} and Table 2, we can observe that:
\begin{itemize}[topsep=0.3ex,parsep=0ex]
% \item The performance of machine learners may decrease due to the redundant features which are generated randomly by Baseline methods.
% \nfo{} solves this problem and achieves improvement on all datasets with an average of $10.65\%$ over Base by decomposing AutoFE into two steps.
\item \nfo{} achieves the best performance in all but four cases.
Although NFS greatly outperforms the baseline methods, \nfo{} still achieves an average improvement of $2.57\%$ over NFS.
For regression tasks, \nfo{} can even achieve a maximum improvement of $11.42\%$.
\item \nfo{} can handle datasets with various numbers of instances and features for both regression and classification tasks and achieve performance improvement on all datasets with an average of $10.72\%$ over Raw and an average of $9.55\%$ over Random.
% the computational cost of a NE algorithm increases proportionally to the network size.
\item 
%Even with an effective searching algorithm, NFS still reduces the feature space to mitigate the feature explosion.
With the benefit of searching in the continuous vector space, \nfo{} addresses the feature explosion problem while preserving the entire space, and achieves highly competitive performance even on large datasets such as \textit{Credit Default} ($30000 \times 25$) and \textit{AP-omentum-ovary} ($275 \times 10936$).
% Table \ref{tab:rq1} shows that \nfo{} is able to handle different datasets as a performance-guided solution to automated feature engineering.
\end{itemize}
\paragraph{Effectiveness of the predictor $\psi_p$} Since the accuracy of the predictor determines the quality of the optimized features, here we demonstrate the effectiveness of $\psi_p$.
We train the feature optimizer using the data augmentation technique mentioned in Section~\ref{s:feat-repr} on an initialized population of 512 features.
After convergence, the loss $\mathcal{L}_{pp}$ (i.e., Mean-Squared Error) of the predictor in the training set is $0.00106$.
To test the predictor, we randomly sample 256 features from the feature space as the test set, which is different from the training set.
%, and use as the metric to measure the performance of $\psi_p$.
The test loss of $\psi_p$ is $0.00132$. Both the training loss and the test loss are small and close, demonstrating the effectiveness of the predictor.
Furthermore, we employ the pairwise accuracy metric to evaluate $\psi_p$.
Let $X$ denote the test set.
$f(x)$ and $y$ denote the predicted performance of $\psi_p$ and the real performance of the feature.
The pairwise accuracy is defined as follows:

\begin{equation}
pairwise\ accuracy = \frac{\sum_{x_{1} \in X, x_{2} \in X}{\mathbb{I}}_{f\left(x_{1}\right) \geq f\left(x_{2}\right)} \mathbb{I}_{y_{1} \geq y_{2}}}{|X|(|X|-1) / 2}
\end{equation}
where $\mathbb{I}$ represents the 0-1 indicator function.
The pairwise accuracy of $\psi_p$ is 0.918, which is close to the ideal value (i.e., 1) and much better than random guess (i.e., 0.5).
% In summary, the differentiable method \nfo{} is shown to be effective and outperforms the existing AutoFE methods.

\begin{figure}
    \captionsetup{labelformat=empty}
    \begin{subfigure}[h]{0.5\linewidth}
        \captionsetup{labelformat=empty}
        \centering
        \parbox[][0.8\textwidth][c]{\linewidth}{
            \centering
            \begin{tabular}{ c | c |  r r }
        		\toprule
        		Dataset&   LFE$^{\ast}$ & NFS$^{\dag}$&     \nfo{}  \\
        		\midrule
        		Credit-a &  0.771& 0.8652& \textbf{0.8826} \\
        		Feritility&  0.873& 0.8700& \textbf{0.9098}  \\
        		Hepatitis&    0.831& 0.8774& \textbf{0.8839}  \\
        		Ionosphere&    0.932& 0.95160& \textbf{0.9770} \\
        		Megawatt1&     0.894& 0.9130& \textbf{0.9171} \\
        		SpamBase& \textbf{0.947}& 0.9296& 0.9339 \\
        		\bottomrule
        	\end{tabular}
        }
    	\caption{Table 2: Comparison between \nfo{}, LFE, and NFS ($^{\ast}$ the results reported in the paper).$\\$}
    \label{tab:comp2}
    \end{subfigure}
    \hfill
    \begin{subfigure}[h]{0.5\linewidth}
        \captionsetup{labelformat=empty}
        \centering
        \includegraphics[width=\textwidth]{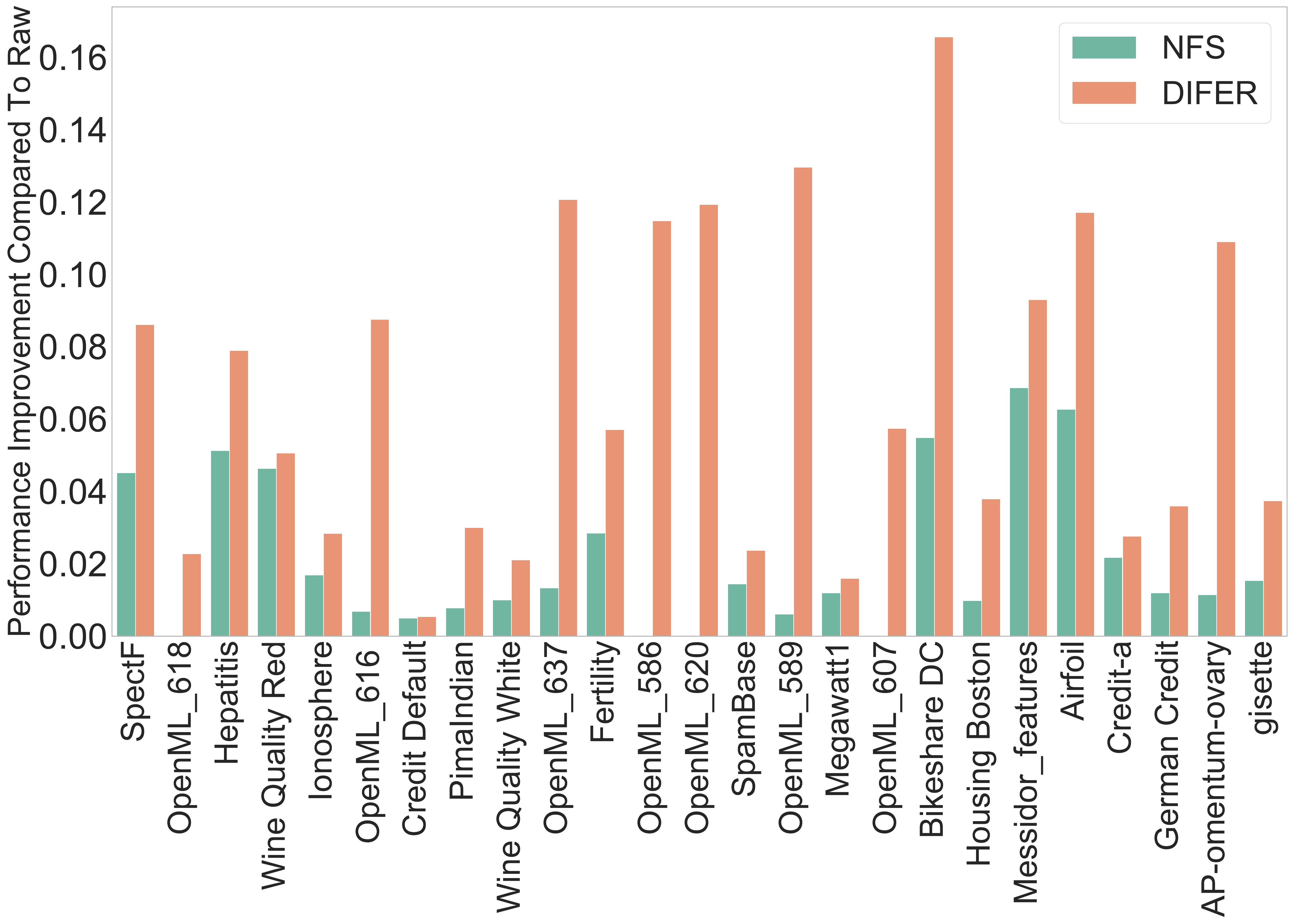}
        \caption{Figure 4: Comparison between NFS and \nfo{}. The number of feature evaluations is restricted to 3500.}
        \label{fig:rq2}
    \end{subfigure}
    % \caption{Comparison between \nfo{}, LFE and NFS.}
    \vspace{-2ex}
\end{figure}

\subsection{Efficiency of \nfo{} (RQ2)}

The overhead of AutoFE can be divided into two parts: the process of feature evaluation and the training overhead of the controller (i.e., the feature optimizer).
To verify the efficiency of \nfo{}, we conduct experiments in terms of the total runtime and the number of feature evaluations, respectively.
Table~\ref{tab:comp1}, where the datasets are sorted in ascending order of model evaluation time, shows the total runtime $\mathcal{T}$ and the average number of feature evaluations for AutoFE, and Figure 4 shows the comparison results between NFS and \nfo{} with a restricted number of feature evaluations.
From Table~\ref{tab:comp1} and Figure 4, we can observe that:
\begin{itemize}[topsep=0.3ex,parsep=0ex]
    \item In Table~\ref{tab:comp1}, \nfo{} achieves better performance than NFS by using 40 times fewer feature evaluations while still achieving significant performance improvement.
    \item From the perspective of runtime, the overhead of \nfo{} is mainly in the training and inference of the feature optimizer compared to NFS which is dominated by feature evaluation.
    Therefore, the efficiency advantage of DIFER is more obvious on larger datasets that requires more evaluation time. For example, compared with NFS, \nfo{} can achieve 2.9$\times$, 7.7$\times$, 11.5$\times$ speedup on \textit{AP-omentum-ovary}, \textit{Credit Default}, \textit{gisette}, respectively.
    \item The advantage of \nfo{} is more significant with a restricted number of feature evaluations measured by Wilcoxon signed-rank test with $p$-value $<0.05$. \nfo{} achieves an average performance improvement of $6.89\%$, doubling that in RQ1.
\end{itemize}

\begin{figure}
    \begin{subfigure}[t]{0.5\linewidth}
        \centering
        \includegraphics[width=\textwidth]{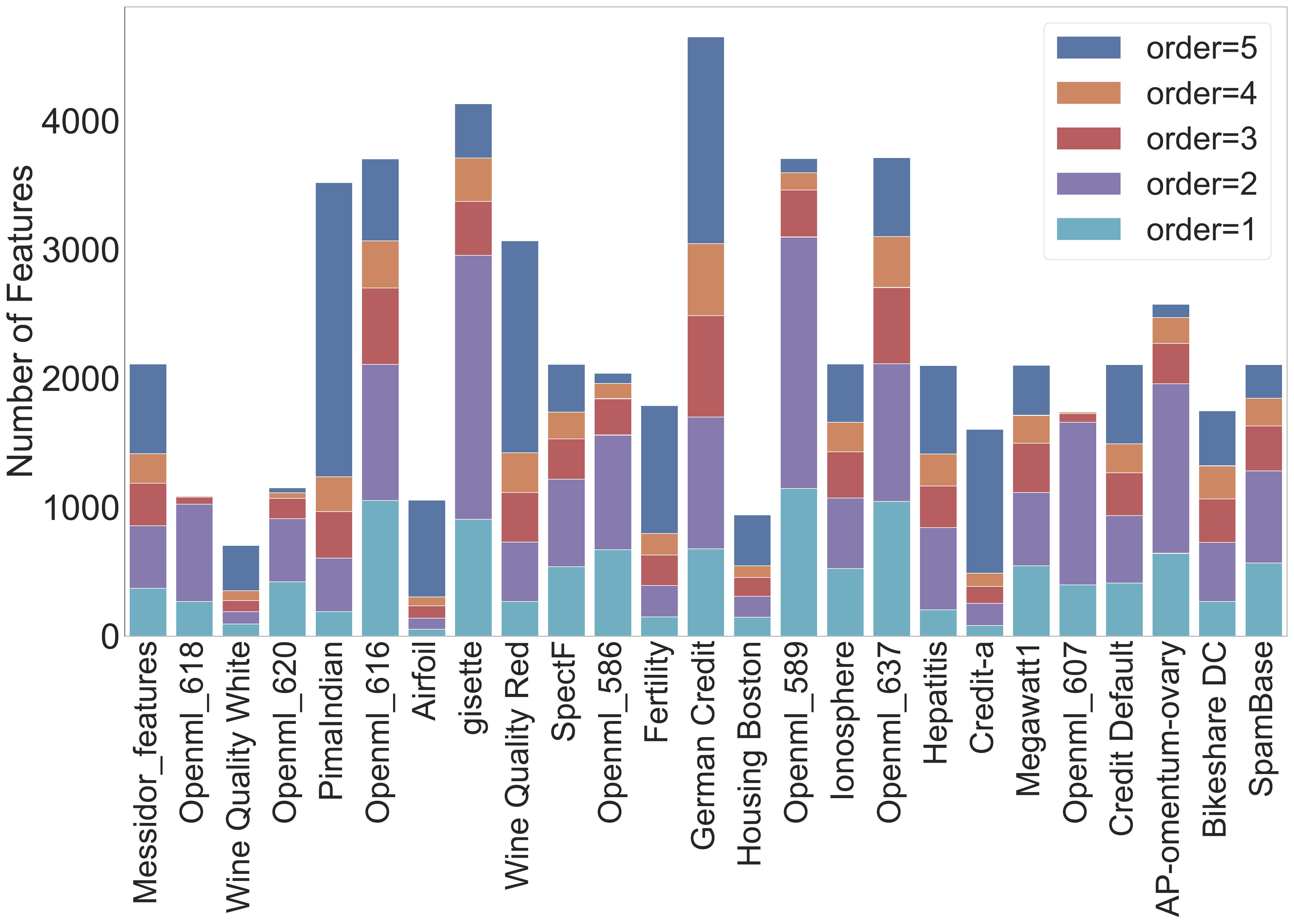}
        \caption{Proportion of different order features generated by \nfo{}.}
        \label{fig:rq3-1}
    \end{subfigure}
    \begin{subfigure}[t]{0.5\linewidth}
        \centering
        \includegraphics[width=\textwidth]{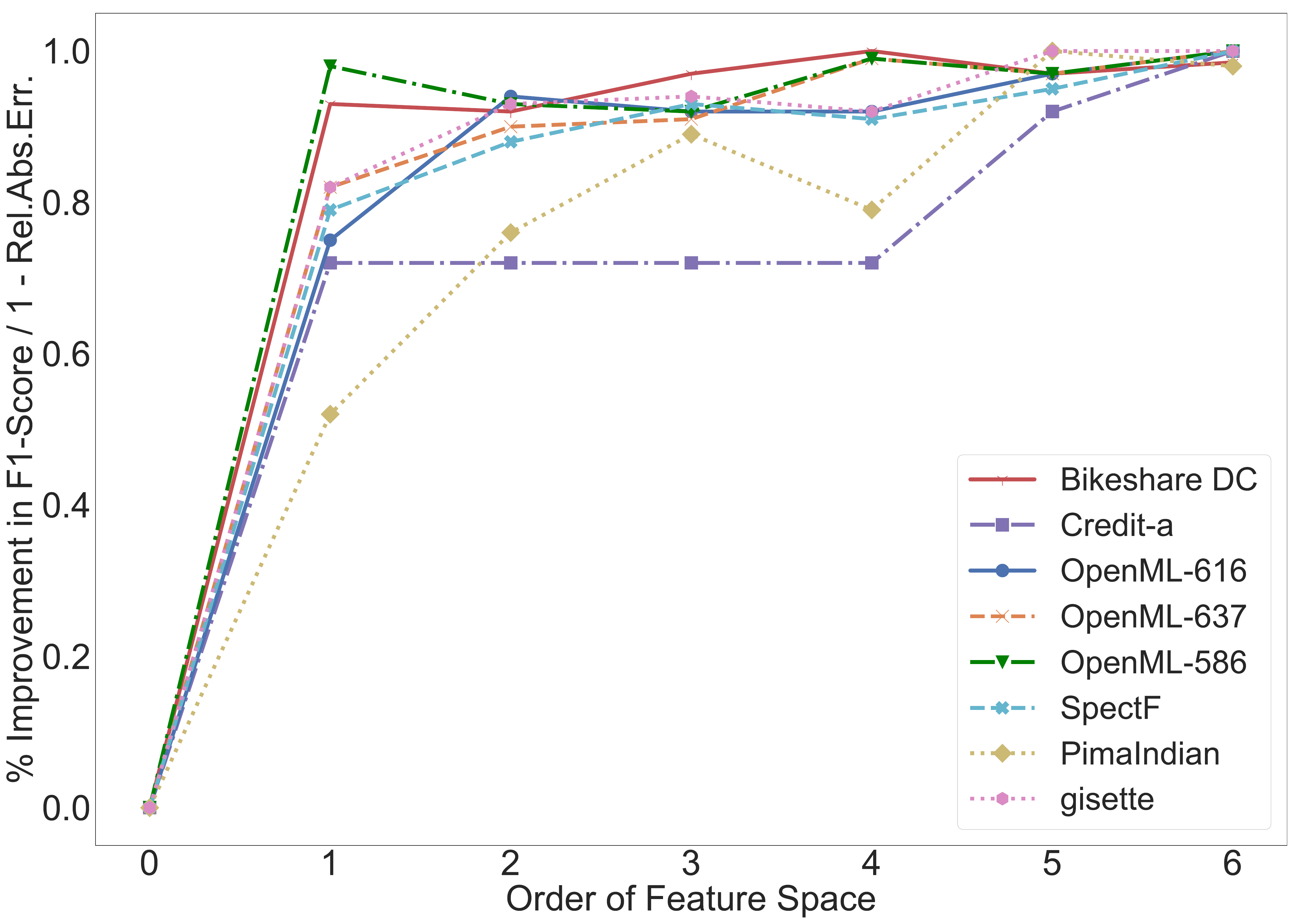}
        \caption{Effect of the high-order feature space.}
        \label{fig:rq3-2}
    \end{subfigure}
      \vspace{-2ex}
    \caption{Effectiveness of high-order features.}
    \vspace{-3ex}
\end{figure}

\subsection{Effectiveness of High-Order Features (RQ3)}

To evaluate the ability of exploring the high-order feature space, we conduct two experiments:

% \begin{figure}[t]
% \centering
% \includegraphics[width=0.43\textwidth]{figures/rq3-1.pdf}
% \vspace{-2ex}
% \caption{Proportion of low/high order features generated by \nfo{}.}
% \label{fig:rq3-1}
% \vspace{-2ex}
% \end{figure}

\begin{enumerate}[topsep=0.3ex,parsep=0ex]
\item Analyze the features generated by \nfo{} during the search process and investigate whether \nfo{} can indeed search for the high-order features.
\item Choose the max order $k$ from $0$ to $6$, where $k=0$ means the raw dataset without any feature transformation. Then, we analyze the performance curve by varying $k$.
\end{enumerate}

Figure \ref{fig:rq3-1} shows the number of each order features generated by \nfo{} with $k=5$ for each dataset. 
High-order features take a considerable average proportion of $80.9\%$, confirming that \nfo{} exploits the entire feature space $F_k^T$ instead of its subspace $F_{i}^T$ where $i < k$.

Besides, we randomly choose 8 datasets, normalize the performance of \nfo{}, plot the performance curve with the increasing max order in Figure \ref{fig:rq3-2}, and draw the following conclusions:

\begin{itemize}[topsep=0.3ex,parsep=0ex]
\item The overall performance of \nfo{} stably increases with the max order $k$. However, when $k$ increases to 5, performance improvement become insignificant.
\item For most datasets, sufficient performance improvement can be already achieved with $k=2$. %Considering that the complexity of extracting features increases linearly with the order, 
There is no need to set an excessively large max order in practice.
\end{itemize}

% The experimental results confirm the effectiveness of \nfo{} in exploring the high-order feature space.
% In other words, \nfo{} can exploit the entire feature space and increase the performance with the max order.

\subsection{Different Machine Learning Algorithms (RQ4)}

To further investigate the generalization ability of \nfo{}, we utilize the commonly-used classification and regression algorithms.
% , including: 
% \begin{itemize}[topsep=0.3ex,parsep=0ex]
% \item LogisticRegression, LinearSVC, XGBoost, and LightGBM~\citep{ke2017lightgbm} for classification.
% \item LassoRegression, LinearSVR~\citep{smola2004svr}, XGBoost~\citep{chen2016xgboost}, and LightGBM for regression.
% \end{itemize}
We conduct experiments on all datasets and the performance statistics are shown in Table \ref{tab:diff-alg}.
Compared to the Raw method, \nfo{} achieves significant improvement under different algorithms.
For instance, LinearSVR with \nfo{} even achieves an average improvement of $32.72\%$ across 25 datasets and a max improvement of $96.98\%$ in \textit{Airfoil}.

% The results show that \nfo{} is independent of machine learning algorithms and achieves performance improvement no matter what algorithm is used.

% The results demonstrate that \nfo{} is independent of machine learning algorithms and can achieve excellent performance improvement with different algorithms.
\setcounter{table}{2}
\begin{table}
    \centering
    \caption{Statistics on the performance of \nfo{} with different ML algorithms.} %The metrics for classification and regression tasks are F1-Score and Relative Absolute Error.
    \scalebox{0.9}{
    \begin{tabular}{c | c | r r}
    \toprule
    Task  &  Algorithm &  Avg Impr$\pm$Std (\%) & Min/Max Impr (\%)\\
    \midrule
    Classification
    & RandomForest &  6.59$\pm$\scriptsize{4.23}  & 0.73 / 15.06 \\
    & LogisticRegression~\citep{hosmer2013logistic} &  5.95$\pm$\scriptsize{4.12} & 1.01 / 15.94 \\
    & LinearSVC~\citep{cortes1995svc} & 13.98$\pm$\scriptsize{9.23} & 3.17 / 22.32 \\
    & XGBoost~\citep{chen2016xgboost} & 6.90$\pm$\scriptsize{6.92}  & 0.30 / 27.98 \\
    & LightGBM~\citep{ke2017lightgbm} & 7.69$\pm$\scriptsize{8.09}  & 0.16 / 32.63 \\
    \hline
    Regression
    & RandomForest &  16.42$\pm$\scriptsize{6.19} & 5.36 / 25.80 \\
    & LassoRegression~\citep{tibshirani1996lasso} & 14.61$\pm$\scriptsize{8.92} & 1.22 / 66.66\\
    & LinearSVR~\citep{smola2004svr} &  32.72$\pm$\scriptsize{19.79} & 13.21 / 96.98\\
    & XGBoost~\citep{chen2016xgboost} & 13.47$\pm$\scriptsize{9.35}  & 3.20 / 67.06 \\
    & LightGBM~\citep{ke2017lightgbm} & 15.46$\pm$\scriptsize{10.48}  & 4.75 / 71.92 \\
    \bottomrule
    \end{tabular}
    }
    
    \label{tab:diff-alg}
    \vspace{-2ex}
    
\end{table}

\section{Conclusion and Future Work}
\label{sec:conclusion}

In this work, we proposed \nfo{}, to the best of our knowledge, the first differentiable AutoFE method. 
\nfo{} leverages an encoder-predictor-decoder-based feature optimizer, which maps features into the continuous vector space via the encoder, optimizes the embedding along the gradient direction induced by the predictor, and recovers better features from the optimized embedding by the decoder.
%%
%The features are first encoded into a continuous vector space.
%%
%Constructing better features can be converted to generating better embedding in the continuous vector space.
%
Moreover, based on the feature optimizer, we proposed a feature evolution framework to search for better features iteratively. 
Experimental results show that \nfo{} is effective on both classification and regression tasks and can outperform the existing AutoFE methods in terms of both prediction performance and computational efficiency.
% In extensive experiments, we achieve quite competitive results on both regression and classification tasks and outperform the existing SOTA in efficiency and performance.

The transformation operations in \nfo{} are for numerical features. For future work, we plan to automatically search for transformations for different feature types (i.e., numerical and categorical).
%We also plan to evaluate \nfo{} on datasets with more input features.

\begin{acknowledgements}
This work was supported in part by the National Natural Science Foundation of China (\#62102177 and \#U1811461), the Natural Science Foundation of Jiangsu Province (\#BK20210181), the Key Research and Development Program of Jiangsu Province (\#BE2021729),  and the Collaborative Innovation Center of Novel Software Technology and Industrialization, Jiangsu, China.
\end{acknowledgements}

\section{Reproducibility Checklist}

All authors must include a section with the AutoML-Conf \textbf{Reproducibility 
Checklist} in their manuscripts, both at submission and camera-ready time.
The reproducibility checklist is a combination of the
\href{https://neurips.cc/Conferences/2021/PaperInformation/PaperChecklist}
     {NeurIPS '21 checklist}
and the
\href{https://www.automl.org/wp-content/uploads/NAS/NAS_checklist.pdf}
     {\textsc{nas} checklist}.
For each question, change the default \verb|\answerTODO{}| (typeset \answerTODO)
to
\verb|\answerYes{[justification]}| (typeset \answerYes),
\verb|\answerNo{[justification]}| (typeset \answerNo), or
\verb|\answerNA{[justification]}| (typeset \answerNA).
\textbf{You must include a brief justification to your answer,} either by
referencing the appropriate section of your paper or providing a brief inline
description.  For example:
\begin{itemize}
\item Did you include the license of the code and datasets? \answerYes{See
  Section~\ref{sec:code}.}
\item Did you include all the code for running experiments? \answerNo{We include
  the code we wrote, but it depends on proprietary libraries for executing on a
  compute cluster and as such will not be runnable without modifications. We
  also include a runnable sequential version of the code that we also report
  experiments in the paper with.}
\item Did you include the license of the datasets? \answerNA{Our experiments
  were conducted on publicly available datasets and we have not introduced new
  datasets.}
\end{itemize}
Please note that if you answer a question with \verb|\answerNo{}|, we expect
that you compensate for it (e.g., if you cannot provide the full evaluation
code, you should at least provide code for a minimal reproduction of the main
insights of your paper).

Please do not modify the questions and only use the provided macros for your
answers. Note that this section does not count towards the page limit. In your
paper, please delete this instructions block and only keep the Checklist section
heading above along with the questions/answers below.
\begin{enumerate}
\item For all authors\dots
  \begin{enumerate}
  \item Do the main claims made in the abstract and introduction accurately
    reflect the paper's contributions and scope?
    \answerYes{}
  \item Did you describe the limitations of your work?
    \answerYes{We discuss its limitations in Section~\ref{sec:conclusion} for further work.}
  \item Did you discuss any potential negative societal impacts of your work?
    \answerNo{}
  \item Have you read the ethics review guidelines and ensured that your paper
    conforms to them?
    \answerYes{}
  \end{enumerate}
\item If you are including theoretical results\dots
  \begin{enumerate}
  \item Did you state the full set of assumptions of all theoretical results?
    \answerNA{}
  \item Did you include complete proofs of all theoretical results?
    \answerNA{}
  \end{enumerate}
\item If you ran experiments\dots
  \begin{enumerate}
  \item Did you include the code, data, and instructions needed to reproduce the
    main experimental results, including all requirements (e.g.,
    \texttt{requirements.txt} with explicit version), an instructive
    \texttt{README} with installation, and execution commands (either in the
    supplemental material or as a \textsc{url})?
    \answerYes{We use open-source datasets and provide source code to reproduce the results in supplemental material.}
  \item Did you include the raw results of running the given instructions on the
    given code and data?
    \answerYes{}
  \item Did you include scripts and commands that can be used to generate the
    figures and tables in your paper based on the raw results of the code, data,
    and instructions given?
    \answerYes{}
  \item Did you ensure sufficient code quality such that your code can be safely
    executed and the code is properly documented?
    \answerYes{}
  \item Did you specify all the training details (e.g., data splits,
    pre-processing, search spaces, fixed hyperparameter settings, and how they
    were chosen)?
    \answerYes{}
  \item Did you ensure that you compared different methods (including your own)
    exactly on the same benchmarks, including the same datasets, search space,
    code for training and hyperparameters for that code?
    \answerYes{See Section~\ref{sec:experiment-setting}.}
  \item Did you run ablation studies to assess the impact of different
    components of your approach?
    \answerYes{}
  \item Did you use the same evaluation protocol for the methods being compared?
    \answerYes{}
  \item Did you compare performance over time?
    \answerNo{}
  \item Did you perform multiple runs of your experiments and report random seeds?
    \answerNo{We use 5-fold cross-validation to reduce randomness.}
  \item Did you report error bars (e.g., with respect to the random seed after
    running experiments multiple times)?
    \answerNo{We use 5-fold cross-validation to reduce randomness.}
  \item Did you use tabular or surrogate benchmarks for in-depth evaluations?
    \answerYes{}
  \item Did you include the total amount of compute and the type of resources
    used (e.g., type of \textsc{gpu}s, internal cluster, or cloud provider)?
    \answerYes{Also see Section~\ref{sec:experiment-setting}.}
  \item Did you report how you tuned hyperparameters, and what time and
    resources this required (if they were not automatically tuned by your AutoML
    method, e.g. in a \textsc{nas} approach; and also hyperparameters of your
    own method)?
    \answerYes{}
  \end{enumerate}
\item If you are using existing assets (e.g., code, data, models) or
  curating/releasing new assets\dots
  \begin{enumerate}
  \item If your work uses existing assets, did you cite the creators?
    \answerYes{We use open-source data with citations.}
  \item Did you mention the license of the assets?
    \answerNo{}
  \item Did you include any new assets either in the supplemental material or as
    a \textsc{url}?
    \answerNo{}
  \item Did you discuss whether and how consent was obtained from people whose
    data you're using/curating?
    \answerNo{We use open-source data and follow protocols.}
  \item Did you discuss whether the data you are using/curating contains
    personally identifiable information or offensive content?
    \answerNo{}
  \end{enumerate}
\item If you used crowdsourcing or conducted research with human subjects\dots
  \begin{enumerate}
  \item Did you include the full text of instructions given to participants and
    screenshots, if applicable?
    \answerNA{}
  \item Did you describe any potential participant risks, with links to
    Institutional Review Board (\textsc{irb}) approvals, if applicable?
    \answerNA{}
  \item Did you include the estimated hourly wage paid to participants and the
    total amount spent on participant compensation?
    \answerNA{}
  \end{enumerate}
\end{enumerate}

% print bibliography -- for bibtex / natbib, use:

\bibliography{automl-conf}

% and for biber / biblatex, use:

% \printbibliography

% supplemental material -- everything hereafter will be suppressed during
% submission time if the hidesupplement option is provided!

\newpage

\appendix

\section{Translation Between Three Forms of Features}
\label{s:bnf}

As mentioned in Section~\ref{s:feat-repr}, there are three forms of features (i.e., original form, parse tree form and traversal string form).
To generate parse trees from the original features, we design the following context-free grammar in BNF (Backus Normal Form):
\begin{itemize}
    \item \textit{ParseTree} $\coloneqq$ $f_{1, \ldots ,d}$ | \textit{UnaryOp ParseTree} | \textit{BinaryOp ParseTree ParseTree}
    \item \textit{UnaryOp} $\coloneqq$ \textit{logarithm} | \textit{abs} | \textit{root} | \textit{min-max} | \textit{normalization} | \textit{reciprocal}
    \item \textit{BinaryOp} $\coloneqq$ \textit{addition} | \textit{subtraction} | \textit{multiplication} | \textit{division} | \textit{modulo}
\end{itemize}
Through such syntax parser, the features are parsed into the form of parse tree, and then the corresponding strings is derived through post-order traversal.
Similarly, due to the many-to-one relationship between traversing strings and parse trees, strings can be reduced to parse trees.
With features as leaf nodes, the constructed features are finally obtained from the bottom up at the root node through the internal nodes with the operators.

\iffalse
Additionally, we show the total number of features added into the original dataset by \nfo{} (i.e., $\lvert \hat{F} \rvert$) in Table~\ref{tab:comp1} and Table 2.
%
% At the end of feature evolution, we select top-$d$ features to form $F_{cand}$. 
%
At the end of feature evolution, we first select top-$d$ features sorted by performance score from $F_{cand}$.
%
Then, we add them one by one to the original dataset with an early-stopping mechanism. 
%
Specifically, we stop adding features when the model performance no longer increases with a patience of 3.
% to the original dataset. Moreover, the maximum value of $k$ is \emph{min($d$, the total number of original features)}.
% Through experiments on various datasets and comparison with existing methods, \nfo{} is shown to be effective and outperform the existing AutoFE methods. 
\fi

\section{Neighborhood of Constructed Features}

The duplicated post-order traversal strings $X = \lbrace x^{(1)}, x^{(2)}, \cdots x^{(n)} \rbrace$ of the feature $\hat{f}_x$ are highly similar in the continuous space:
\begin{equation}
    \exists \epsilon,\ \forall x^{(i)} \in X, \Vert e_{x^{(i)}} - \frac{1}{n} \sum_{x^{(j)} \in X} e_{x^{(j)}} \Vert_2 \leq \epsilon
    \label{eq:neighbour}
\end{equation}
where $\epsilon$ is just a variable used to inscribe the property, not a hyperparameter.
The neighborhood of $X$ in the continuous space can be represented as $\delta_X=\lbrace e_{x^{'}} |\Vert e_{x^{'}} - \frac{1}{n} \sum_{x^{(j)} \in X} e_{x^{(j)}} \Vert_2 \leq \epsilon \rbrace$.
After one step of gradient ascent, the decoded string of $e_{x^{'}} \in \delta_x$ may still be in $X$.
% As shown in Figure \ref{fig:emb-space}, $e_{x^{'}}$ may still be in the neighborhood of $X$ after gradient ascent.
Thus, the same parse tree is got.
To escape the neighborhood., we use multi-step gradient ascent as mentioned in Section 3.5.

\section{Adaptive Loss Weight Setting}
\label{s:adaptive-loss}
We use the parameter $\lambda \in \mathcal{R}^{+}$ to balance $\mathcal{L}_{pp}$ and $\mathcal{L}_{rec}$ and $\lambda$ is determined adaptively. Inspired by~\citep{goyal2017warmup}, the first $k$ epochs are used to warm up the jointly-training of the feature optimizer with $\lambda = 1$. 
After the first $k$ epochs, we assign $\lambda = \sum_{i=1}^{k} \mathcal{L}_{rec} / \sum_{i=1}^{k} \mathcal{L}_{pp}$ according to the sum of losses. This is mainly to make the two losses in the same order of magnitude.
In practice, $k$ is empirically set to 5.

\section{Robustness}
\label{s:hyper}

\begin{figure}[htp]
    \begin{subfigure}[t]{0.48\linewidth}
        \centering
        \includegraphics[width=\textwidth]{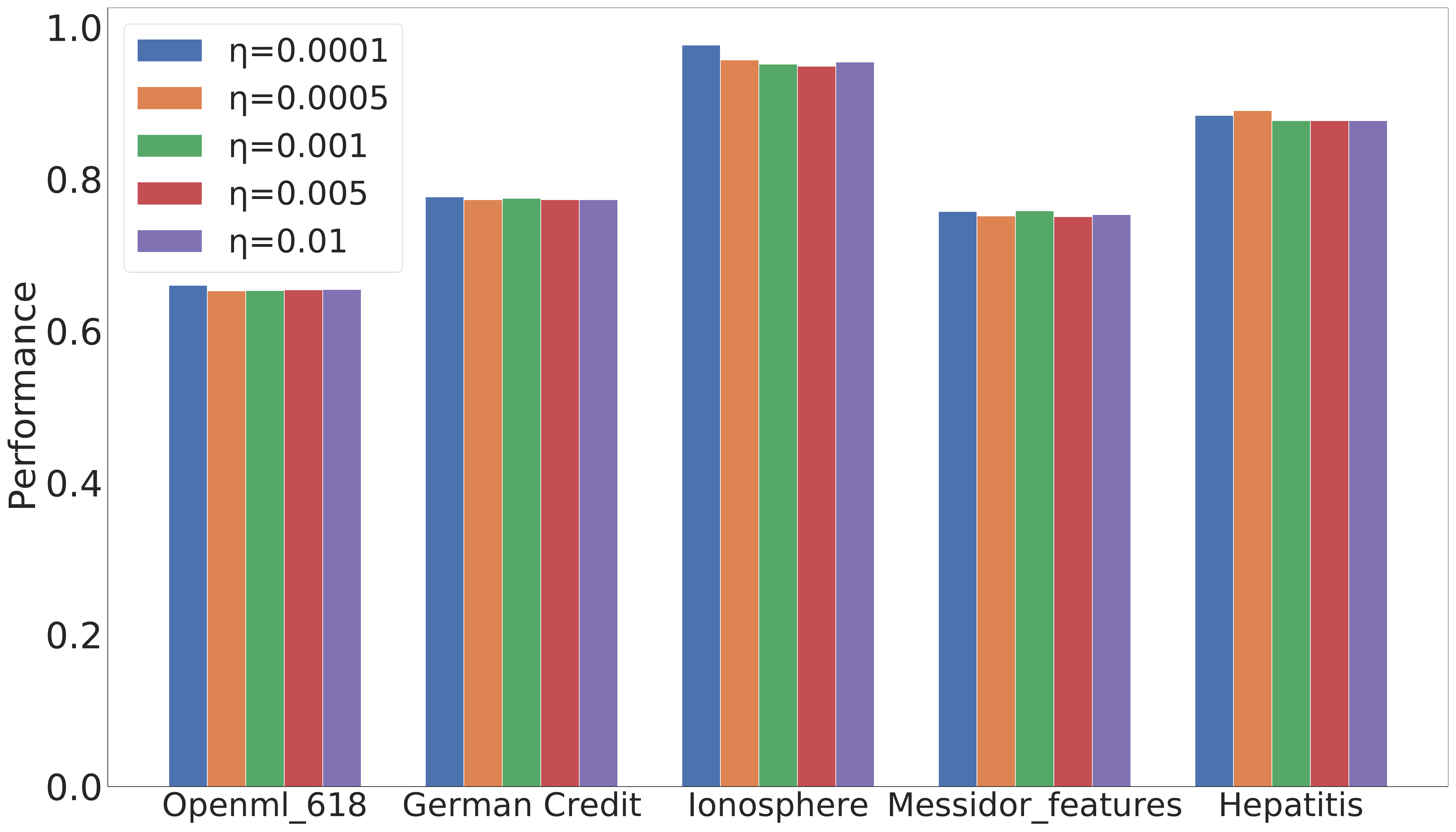}
        \caption{Comparison results of \nfo{} with different settings of the evolution rate $\eta$.}
        \label{fig:rq5-evo}
    \end{subfigure}
    \begin{subfigure}[t]{0.48\linewidth}
        \includegraphics[width=\textwidth]{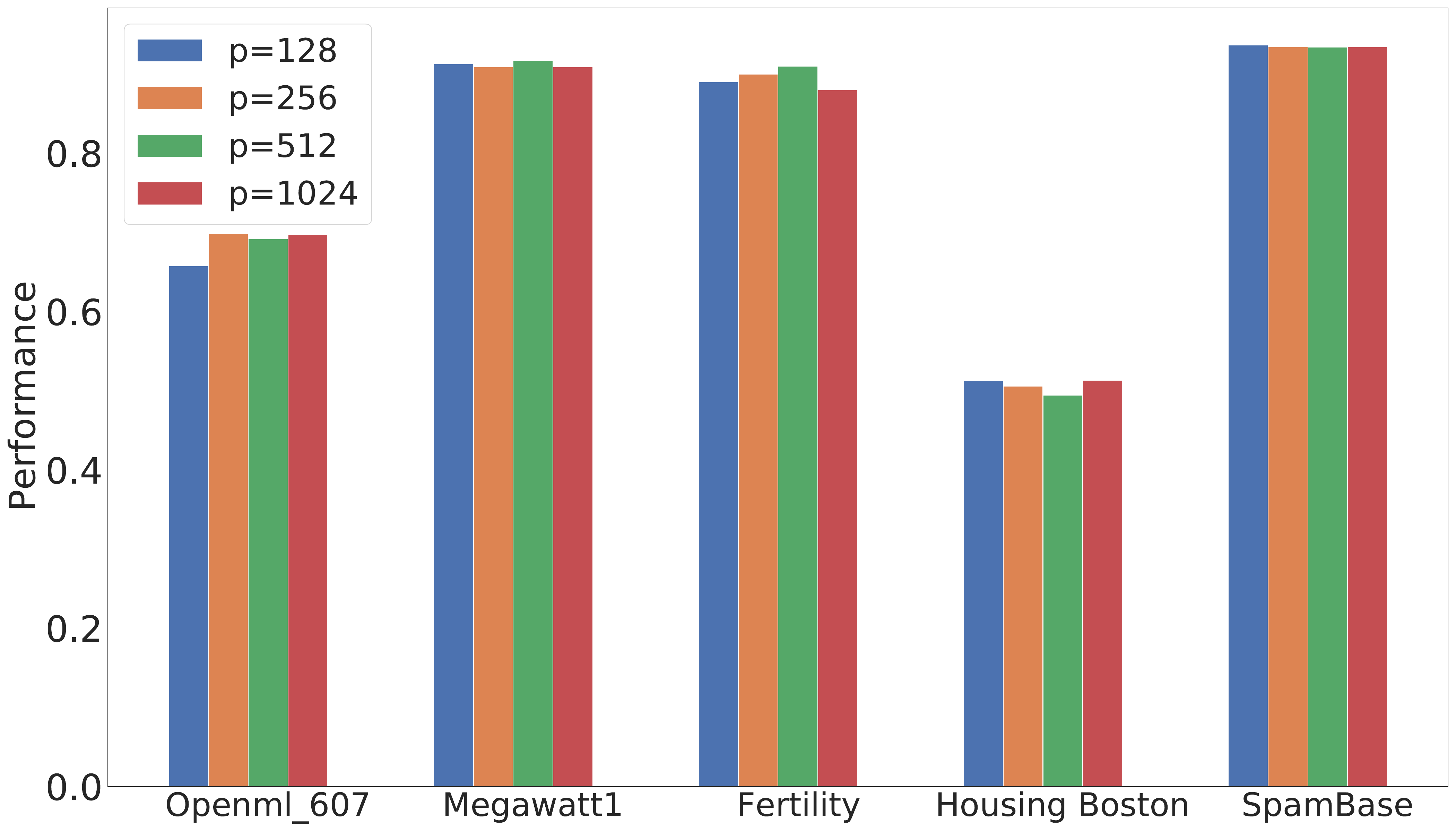}
        \vspace{-2ex}
        \caption{Comparison results of \nfo{} with different settings of the population size $p$.}
        \label{fig:rq5-psize}
    \end{subfigure}
    \caption{Robustness of \nfo{}.}
\end{figure}

% We perform the experiments on the same datasets used in RQ2.
We further evaluate whether \nfo{} is sensitive to different hyperparameters, including evolution rate $\eta$ and population size $p$.
Figure \ref{fig:rq5-evo} and Figure \ref{fig:rq5-psize} show the experimental results on 5 randomly-selected datasets that represent both classification and regression tasks.
Figure~\ref{fig:rq5-evo} demonstrates that \nfo{} is robust to different settings of $\eta$.
Empirically, $\eta$ should not be too large in gradient ascent.
A small $\eta$ can get the same or even better results than the large $\eta$ by performing multiple times of gradient ascent.
Moreover, a larger $p$ allows the feature optimizer to be fully trained, and a smaller $p$ allows more features to be optimized in the case of a limited number of feature evaluations.
As shown in Figure \ref{fig:rq5-psize}, the performance of \nfo{} remains stable across different settings of $p$.  

\section{Statistics Comparison}

To further statistically evaluate the difference between the AutoFE methods in Table \ref{tab:comp1}, we perform the Friedman test~\cite{JMLR:v7:demsar06a}, which is a non-parametric equivalent of the repeated-measures ANOVA.
It is used to determine whether or not there is a statistically significant difference.

For the comparison results in Table~\ref{tab:comp1}, we first calculate the Friedman statistic. 
Let $r_i^j$ be the rank of the $j$-th of $k$ AutoFE methods ($k$ = 4, i.e., DFS, AutoFeat, NFS, and \nfo) on the $i$-th of $N$ datasets. 
The Friedman test compares the average ranks of models, $R_j = \frac{1}{N}\sum_i{r_i^j}$.
The null-hypothesis states that all the tree models are equivalent and so their ranks $R_j$ should be equal.
We employ the scipy tool\footnote{https://github.com/scipy/scipy} to calculate the Friedman statistic.
The corresponding Friedman $p$-value is 1.17e-10.
Since the $p$-value is less than 0.05, we can reject the null hypothesis that the performance is the same for all four types of AutoFE methods.
In other words, we have sufficient evidence to conclude that the AutoFE method lead to statistically significant differences in terms of performance.
Since the $p$-value of the Friedman test is statistically significant, we perform the Nemenyi post-hoc test~\cite{nemenyi1963distribution} to further determine exactly which AutoFE method has different means.
Table~\ref{tab:stats} shows the $p$-values for each pairwise comparison.
We can conclude that \nfo{} is significantly different from other trees for a confidence level of $p=0.05$ and show the result by '*' in Table\ref{tab:comp1}.

\begin{table}[]
\centering
\caption{$p$-values for each pairwise comparison using the Nemenyi post-hoc test for the AutoFE methods (Confidence level $p=0.05$).}
\begin{tabular}{@{}lllll@{}}
\toprule
          & DFS    & AutoFeat   &  NFS      & \nfo{}      \\ \midrule
DFS      & 1.0000  & 0.2967    & 0.0010   & \textbf{0.0010} \\
AutoFeat & 0.2967  & 1.0000    & 0.0313 & \textbf{0.0010}   \\
NFS       & 0.0010  & 0.0313  & 1.0000   & \textbf{0.0046} \\
\nfo{}       & \textbf{0.0010} & \textbf{0.0010}    & \textbf{0.0046} & 1.0000   \\ \bottomrule
\end{tabular}
\label{tab:stats}
\end{table}

\section{Numbers of Added Features}

In the formal definition, $|F^{*}|$is defined as the set of added features, which does not contain the raw features.
Table \ref{tab:num-of-feat} shows the number of features finally added by AutoFeat, NFS and \nfo{}.
For NFS, it is the number of original features.
Benefiting from feature selection, $|F^{*}|$ in AutoFeat and \nfo{} is adaptive.

\begin{table*}[t]
	\centering
    \caption{The number of features $|F^{*}|$ added into the original dataset by AutoFE methods (\textit{Err.} indicates failure due to out of memory when running the open-source code).}

	\scalebox{1.0}{
		\begin{tabular}{ c | c c | r r r}
			\toprule
			Dataset&   C/R&    Inst.$\backslash$Feat.& $\lvert F^{*} \rvert_{AutoFeat}$  & $\lvert F^{*} \rvert_{NFS}$  & $\lvert F^{*} \rvert_{\nfo{}}$ \\
			\midrule

            Housing Boston&    R&  506$\backslash$13&  21 &  13 & 1 \\

			Bikeshare DC&  R&  10886$\backslash$11& 17 & 11 & 6 \\

			Airfoil&  R&  1503$\backslash$5& 4 & 5 & 4 \\
			
			Openml\_586&    R&  1000$\backslash$25& 37  & 25 & 20 \\
			
			Openml\_589&    R&  1000$\backslash$25& 21 & 25 & 20 \\
			
			Openml\_637&    R&  1000$\backslash$25& 30 & 25 & 13 \\

			Openml\_618&    R&  1000$\backslash$50& 49 & 50 & 32 \\

			Openml\_607&    R&  1000$\backslash$50& 51 & 50 & 38 \\
			
			Openml\_616&    R&  1000$\backslash$50& 41 & 50 & 8 \\

			Openml\_620&    R&  1000$\backslash$50& 32 & 50 & 12 \\

			\midrule
			
			Hepatitis&      C&  155$\backslash$6& 7 & 6 & 6 \\
			
			Fertility&     C&  100$\backslash$9& 12 & 9 & 3 \\

			SpectF&    C&  267$\backslash$44&  37 & 44 & 9 \\

			Megawatt1& C&  253$\backslash$37& 48 & 37 & 29\\

			Ionosphere&     C&  351$\backslash$34& 52 & 34 & 1 \\

			German Credit&     C&  1001$\backslash$24& 22 & 24 & 1 \\

			Credit-a&    C&  690$\backslash$6& 4 & 6 & 5 \\

			PimaIndian&  C&  768$\backslash$8&  12 & 8& 1 \\

			Messidor\_features&    C&  1150$\backslash$19& 29 & 19 & 10\\

			Wine Quality Red&  C&  999$\backslash$12& 8 & 12 & 6 \\
			
			Wine Quality White&   C&  4900$\backslash$12& 11 & 12 & 9 \\
			
			SpamBase&   C&  4601$\backslash$57& 46 & 57 & 1 \\
			
			AP-omentum-ovary & C& 275$\backslash$10936 & \textit{Err.} & 10936 & 491 \\
			
			Credit Default&    C&  30000$\backslash$25& 30 & 25 & 5 \\

			gisette & C& 2100$\backslash$5000 & \textit{Err.} & 5000 & 19 \\
			
			\bottomrule
		\end{tabular}

	}
	\label{tab:num-of-feat}
	 \vspace{-3ex}
\end{table*}

\end{document}

\appendix

\section{Details of Experimental Settings}

\subsection{Algorithm Details}

\paragraph{Population Initialization Step.}
In this phase, we randomly sample 512 features from the feature space.
Specifically, we randomly select transformations for the raw features.
The detailed process of random sampling can be viewed in the \textit{feat\_tree.py}.
We use the cross-validation evaluation method and divide the dataset $D_{\text{origin}} \cup \{f\}$ into 5 folds.
For each fold, we use it as a validation set and train a machine learning model from scratch using the remaining data.
The mean of the trained model's performance metric on the validation set is used as the performance score $s$ of the feature.

\iffalse
\paragraph{Optimizer Training.} The goal of this phase is to train a feature optimizer with the $\langle \textit{Feature\_Str}, \textit{Score}\rangle$ set.
We train the feature optimizer with $400$ epochs using the Adam with a learning rate of $0.001$ and a weight decay of $0.0001$, where the batch size is $128$.
To avoid overfitting, we use EarlyStop with patience of $10$.
\fi

\paragraph{Feature Evolution Step.} In this core phase, we evolve the randomly sampled features to improve their performance.
First, we select the top-$d$ features based on their performance scores.
The value of d is empirically set to be the minimum between top 20\% of the initial population size and the total number of original features.
Next, the top features are turned into traversal strings as the input of the encoder.
Then, we employ the encoder to encode the string into a continuous embedding, perform gradient descent multiple times with a learning rate of $0.0001$ along the score direction predicted by the predictor, and decode the optimized embedding by the decoder to get the traversal strings.
Finally, the optimized traversal strings are recovered to features according to the parsing rules in \textit{feat\_tree.py}.

\subsection{Compared Methods}

We compare \nfo{} with the following state-of-the-art and baseline methods:

\begin{enumerate}
    \item \textbf{Base} is a baseline method, which directly use the original dataset to train the machine learning model from scratch.
    \item \textbf{Random} is a baseline method, which randomly samples features from the feature space and adds the features to the original dataset.
    \item \textbf{DFS}\cite{kanter2015deep} is a well-known approach based on \textit{expansion-reduction}. All transformations are applied separately and added to the raw features, followed by a feature selection routine.
    The code is available on \url{https://github.com/alteryx/featuretools}, so we can run DFS directly and obtain the experimental results.
    \item \textbf{AutoFeat}\cite{horn2019autofeat} is a state-of-the-art approach, which iteratively subsamples features using beam search.
    It is open-source on \url{https://github.com/cod3licious/autofeat}, so we can run AutoFeat on the datasets to get the experimental results.
    \item \textbf{LFE}\cite{nargesian2017learning} recommends the most promising transformation for each feature. There is a set of MLP classifiers, each corresponding to a transformation.
    Note that LFE is limited to the classification tasks.
    Since the code of LFE is not available, we use the experimental results directly from the paper.
    \item \textbf{NFS}\cite{chen2019neural} is the state-of-the-art method, which treats features as strings and utilizes several RNN-based controllers to generate transformation sequences.
    Since some of the experimental results in the NFS paper are evaluated using \textit{java.weka} package, we use its open-source code on \url{https://github.com/TjuJianyu/NFS} to obtain the results.
\end{enumerate}

\section{Detailed Results in RQ4}

We list here in detail the results of \nfo{} applied to other machine learning algorithms on various datasets.

Table \ref{tab:regression} shows the comparison results of Base and \nfo{} on the machine learning algorithm LassoRegression, LinearSVR, XGBoost and LightGBM.
Table \ref{tab:classification} shows the comparison results of Base and \nfo{} on the machine learning algorithm LogisticRegression, LinearSVC, XGBoost and LightGBM.

\begin{table*}[htb]
\scalebox{1.0}{
\begin{tabular}{ |c | r r | r r | r r | r r | r r |}
\hline
\multirow{2}*{Dataset}&  \multicolumn{2}{c|}{LassoRegression}&   \multicolumn{2}{c|}{LinearSVR}& \multicolumn{2}{c|}{XGBoost}&     \multicolumn{2}{c|}{LightGBM}\\
\cline{2-9}
& Base & \nfo{} & Base & \nfo{}  & Base & \nfo{}  & Base & \nfo{} \\
\hline
Openml\_586&   -0.0019&  0.0131& 0.1033& 0.3920&  0.7200 & 0.7290 & 0.7370 & 0.7620\\

Openml\_589&  -0.0029&  0.0117& 0.1510& 0.3852&  0.7162 & 0.7280 &0.7465 & 0.7618\\

Openml\_607&   -0.0022&  0.0284& 0.0552& 0.1800&  0.7120 & 0.7231 & 0.7246 & 0.7390\\

Openml\_616&    -0.0007&  0.0319& -0.0024& 0.1955& 0.6397 & 0.6529 & 0.6704 & 0.6870 \\

Openml\_618&   -0.0039&  0.0171& 0.0902&  0.2418& 0.6921 & 0.7102 & 0.7090 & 0.7304\\

Openml\_620&    -0.0027&  0.0095& 0.0812&  0.3821& 0.7043 & 0.7151 & 0.7374 & 0.7499\\

Openml\_637& 0.0010&  0.0402& 0.0039& 0.1455&  0.6177 & 0.6552 & 0.6751 & 0.6939\\

\hline

Bikeshare DC&   0.9999& 0.9999& 0.9999& 0.9999 & 0.8391 & 0.9470 & 0.8233 & 0.9504\\

\hline

Housing Boston&   0.2978&  0.4910& 0.1201& 0.3318& 0.4776 & 0.5210 & 0.4241 & 0.5018 \\

Airfoil& 0.4962&  0.0130& -29.8430 & 0.0681& 0.5118 & 0.6329 & 0.5144 & 0.6402 \\
\hline
\end{tabular}
}
\centering
\caption{Performance of \nfo{} with Different Machine Learning Algorithms on Regression tasks.}
\label{tab:regression}
\end{table*}

\begin{table*}[htb]
\scalebox{1.0}{
\begin{tabular}{| c | r r | r r | r r | r r | r r  |}
\hline
\multirow{2}*{Dataset}&  \multicolumn{2}{c|}{LogisticRegression}&   \multicolumn{2}{c|}{LinearSVC}& \multicolumn{2}{c|}{XGBoost}&     \multicolumn{2}{c|}{LightGBM}\\
\cline{2-9}
& Base & \nfo{} & Base & \nfo{}  & Base & \nfo{}  & Base & \nfo{} \\
\hline
AP-omentum-ovary&   0.7818 &  0.8036& 0.7855& 0.8436&  0.8218 & 0.8564 & 0.8255 & 0.8773\\

PimaIndian&   0.7683 &  0.7852& 0.6238& 0.7540&  0.7527 & 0.7748 & 0.7475 & 0.7839\\

SpectF&   0.7078&  0.8200& 0.7156& 0.8091&  0.8164 & 0.8428 & 0.8014 & 0.8877\\

German Credit&   0.7540&  0.7710& 0.6710& 0.7470&  0.7650 & 0.7970 & 0.7680 & 0.8120\\

Ionosphere&   0.8520&  0.9289& 0.8610& 0.9261&  0.9204 & 0.9602 & 0.9346 & 0.9630\\

Credit Default&  0.7787&  0.7865& 0.0000& 0.0000&  0.8142 & 0.8172 & 0.8205 & 0.8218\\

Messidor\_features &  0.7498&  0.7672& 0.6264& 0.7316&  0.6959 & 0.7698 & 0.6976 & 0.7863\\

Wine Quality Red&   0.5716&  0.5954& 0.4584& 0.5541&  0.5472 & 0.5929 & 0.5570 & 0.5954\\

Wine Quality White&   0.4545&  0.5270& 0.0000& 0.0000&  0.5112 & 0.5400 & 0.5092 & 0.5341\\

SpamBase &   0.8990&  0.9252& 0.0000& 0.0000&  0.9315 & 0.9440 & 0.9318 & 0.9418\\

Credit-a &   0.8000&  0.8610& 0.7015& 0.8580& 0.8348 & 0.8710 & 0.8390 & 0.8725\\

Fertility&   0.8300&  0.9000& 0.8300& 0.9000&  0.8100 & 0.9100 & 0.8200 & 0.9000\\

Hepatitis&   0.7871&  0.8065& 0.7550& 0.8710& 0.8065 & 0.8903 & 0.7871 & 0.8774\\

Megawatt1&   0.8971&  0.9249& 0.8774& 0.9053&  0.8730 & 0.9287 & 0.8656 & 0.9289\\

gisette&   0.9644&  0.9762& 0.9626& 0.9744&  0.9803 & 0.9819 & 0.9707 & 0.9806\\
\hline
\end{tabular}
}
\centering
\caption{Performance of \nfo{} with Different Machine Learning Algorithms on Classification tasks.}
\label{tab:classification}
\end{table*}

\bibliographystyle{named}
\bibliography{ijcai21}